\documentclass{article}

\usepackage{amsmath}
\usepackage[preprint]{neurips_2026}

\usepackage[utf8]{inputenc} 
\usepackage[T1]{fontenc}    
\usepackage{hyperref}       
\usepackage{url}            
\usepackage{booktabs}       
\usepackage{amsfonts}       
\usepackage{amsthm}         
\usepackage{nicefrac}       
\usepackage{microtype}      
\usepackage[table]{xcolor}  
\usepackage{colortbl}
\usepackage{algorithm}
\usepackage{algpseudocode}
\usepackage{multirow}
\usepackage{graphicx}
\usepackage{subcaption}

\usepackage{enumitem}

\usepackage{titletoc}
\usepackage{float}

\usepackage{amssymb}
\usepackage{bm}

\usepackage{array}
\usepackage{tcolorbox}

\usepackage{mathtools}

\usepackage[capitalize,noabbrev]{cleveref}

\algdef{SE}[PARFOR]{ParFor}{EndParFor}[1]{\textbf{for} #1 \textbf{do in parallel}}{\textbf{end for}}
\algtext*{EndParFor}

\theoremstyle{plain}
\newtheorem{theorem}{Theorem}[section]

\newtheorem{lemma}[theorem]{Lemma}

\theoremstyle{definition}

\newtheorem{assumption}[theorem]{Assumption}
\theoremstyle{remark}

\usepackage[textsize=tiny]{todonotes}

\title{TAP: Two-Stage Adaptive Personalization of Multi-Task and Multi-Modal Foundation Models in Federated Learning}

\author{%
  Seohyun Lee \\
  Purdue University\\
  \And
  Wenzhi Fang \\
  Purdue University \\
  \And
  Dong-Jun Han \\
  Yonsei University \\
  \AND
  Seyyedali Hosseinalipour \\
  University at Buffalo-SUNY \\
  \And
  Christopher G. Brinton \\
  Purdue University \\
}

\begin{document}

\maketitle
\newcommand{\E}{\mathbb{E}}
\newcommand{\KL}{\mathrm{KL}}
\newcommand{\softmax}{\mathrm{softmax}}
\newcommand{\tok}{\mathrm{tok}}
\newcommand{\concat}{\mathbin{\|}}

\begin{abstract}
In federated learning (FL), local personalization of models has received significant attention, yet personalized fine-tuning of foundation models remains underexplored. In particular, there is a lack of understanding in the literature on how to personalize foundation models in settings where there exist heterogeneity not only in data, but also in tasks and modalities across the clients. To address this gap, we propose Two-Stage Adaptive Personalization (TAP). In the first stage, TAP leverages mismatched model architectures between clients and the server to selectively replace personalized parameters with global updates, explicitly limiting cross-task and cross-modality interference. In the second stage, TAP conducts post-FL distillation on the global model to recover a beneficial shared structure. By reintroducing generalizable knowledge only after the global model has stabilized, TAP enhances generalization without compromising personalization.
In developing our methodology, we introduce the first convergence analysis of federated foundation model training at the server under modality-task pair heterogeneity across clients, and demonstrate the impact of the number of modality-task pairs on model fine-tuning. Through extensive experiments, we demonstrate the effectiveness of TAP across a variety of datasets and tasks in comparison to state-of-the-art baselines. The implementation code is publicly available \href{https://github.com/lee3296/TAP}{here}.
\end{abstract}

\section{Introduction}

Federated Learning (FL) is a distributed machine learning paradigm featuring the ability to collaboratively train a model without sharing raw data. In FL, each client hosts its own model, trains it locally, and transmits it to a central server for aggregation~\citep{konevcny2016federated, mcmahan2017communication}. In this vein, recent attention has been given to the question of how to train/fine-tune foundation models in a federated setting, with large language models (LLMs)~\citep{fang2025federated, ye2024openfedllm} being the most extensively studied model class. Foundation Models, as outlined by \cite{bommasani2021opportunities}, are models trained on large amounts of data that can be fine-tuned and adapted to a range of downstream tasks~\citep{lian2024less}. In this regard, a large variety of foundation models have been developed and considered, such as DALL-E~\citep{marcus2022very} and the GPT family~\citep{achiam2023gpt}. These base models are often considered \textit{pre-trained}, and fine-tuning of such models have been shown to offer promise across a variety of applications, such as medical imaging~\citep{zhang2025parameter}. 

In general, due to the collaborative nature of FL, the final model produced by the training process may not be particularly well-suited to each individual client. To address this, various techniques for personalized FL (PFL) have been explored~\citep{tan2022towards, deng2020adaptive}. In PFL, the server facilitates training of the collaborative model while also allowing each client to tailor or fine-tune a model to its own local data. 

\textbf{Challenges:} Despite many promising approaches to PFL, the majority of explored applications are limited to \textit{uni-modal, uni-task} scenarios, where the clients and server share the same model architecture, task, and modality of input data. Moreover, emerging works that have studied PFL fine-tuning of larger foundation models \cite{chen2024feddat, luo2024mixture, long2024dual} have made a limiting assumption of a uniform architecture among all clients and the server (even when dealing with multi-modal and/or multi-task models), as it enables use of standard aggregation protocols. \textit{\textbf{In many practical FL scenarios, however, each client's modalities and tasks will differ, motivating use of non-uniform model architectures and necessitating the learning process to account for such heterogeneities.}} For example, in a healthcare setting, differing institutions (e.g., hospitals, clinics) may collect non-identical sets of modalities (e.g., image scans, text reports) and pursue distinct tasks such as diagnosis prediction or report generation. Similarly, in smart manufacturing, factories may have differing sensor systems (e.g. acoustic vs. temperature measurement) that seek to optimize differing task objectives, such as safety or defective product detection. Therefore, a crucial question is: \textit{How to personalize local foundation models at clients when models returned from the server are themselves heterogeneous, as are the tasks and modalities they support?}

\subsection{Contributions}\label{subsec:contributions}
In this work, we develop a novel PFL methodology enabling personalization of foundation models that are heterogeneous in tasks and modalities. The overall division of labor in our scheme is intuitive: the server learns a generalizable model capable of handling all tasks and modalities, while the clients adapt their local foundation models towards their own task/modality compositions. To facilitate this, we propose a two-stage adaptive personalization (TAP) process based on two key ideas. First is \textbf{\textit{precise personalization on an individual modality-task pair basis}}: during FL training, each client will hold personalized parameters in addition to the model transmitted to the server, and will strategically replace a subset of its personalized parameters when it perceives that the returned model from the server will benefit its personal parameters on a specific task. This permits each client to effectively pick-and-choose beneficial modality-task pairings with limited interference among pairs under multi-modal, multi-task conditions. Second is \textbf{\textit{knowledge distillation \cite{jiang2020federated} in the post-FL period}}: the returned model from the server is employed as a teacher, designed to enable incorporation of generalizable knowledge from other modality-task pairings. In contrast with existing works on knowledge distillation in FL, our procedure is carefully designed to \textit{limit cross-task and modality interference}, a challenge non-existent in the conventionally studied homogeneous architectural scenarios \cite{wang2023dafkd}.
In developing our methodology, our key contributions are as follows:

\begin{itemize}[leftmargin=*]
    \item We provide the first convergence analysis of the server model in component-based PFL for heterogeneous modality-task foundation modeling. Our analysis shows that convergence of the system-wide model degrades as the number of modality-task pairs increases. This exposes a central bottleneck in scaling PFL across heterogeneous foundation models and motivates client-side personalization.
    
    \item To accomplish this, we propose the two-stage adaptive personalization (TAP) methodology, where clients intelligently leverage parameters learned in the vanilla FL protocol when deemed beneficial to their local modality-task compositions. In the first stage, leveraging client-defined margin hyperparameters, each client conducts selective parameter replacement when the model engaged in the FL process is inferred to offer benefit to the personalization process. Afterwards, in the second stage, a knowledge distillation-based post-FL process fine-tunes the personalized model by distilling knowledge from the FL-engaged model while maintaining personalization.
    
    \item We conduct extensive experiments across eight datasets encompassing image and text classification and generation tasks, comparing performance of our proposed method with the state-of-the-art baselines in PFL fine-tuning. We show that TAP consistently obtains superior performance metrics, emphasizing that our method is better suited for personalization of foundation models when considering the additional complexity of heterogeneity in modalities and tasks across clients.
\end{itemize}

\section{Related Work}

\textbf{Personalized Federated Learning (PFL): }In PFL, clients aim to train local models that are personalized towards their own data while simultaneously training a global model via the FL training process. In general, personalization in PFL falls into two categories: (i) personalized fine-tuning embedded into global model training~\citep{kairouz2021advances, mansour2020three}, and (ii) training of individual personalized models informed by but separate from the typical FL learning process~\citep{tan2022towards, ghuhan2019federated}. 
Our work is interested in PFL for foundation models specifically, tackling the heterogeneous modality-task challenge. Existing works dealing with multi-modal~\citep{luo2024mixture} and/or multi-task PFL~\citep{chen2024feddat, long2024dual, lu2024fedhca2} implicitly assume that the tasks and/or modalities are homogeneous across all clients, thereby justifying a common model architecture. In particular, while \cite{lu2024fedhca2} explores heterogeneous model architectures in a multi-task setting, it is specifically designed for image inputs and therefore is uni-modal. In this respect, \cite{chen2024disentanglement} proposed to have separate encoders and decoders for differing modalities and tasks, with a shared transformer backbone utilizing a Mixture of Experts (MoE)~\citep{yuksel2012twenty} structure to route inputs based on modality-task pairs. Aiming to promote more separable latent spaces between pairs, \cite{chen2024disentanglement} introduces a disentanglement auxiliary loss, which can nonetheless entangle local models that share common subsets of the server model during aggregation. By contrast, TAP seeks to decouple itself from the FL training process, introducing personalized parameters which only interact with the server model when they signal benefit to the personalization process, avoiding loss of granularity on specific tasks. Our convergence analysis in Sec.~\ref{sec:conv} motivates such a decoupling, which we see in Sec.~\ref{sec:experiments} enables TAP to obtain significant improvements over baselines including \cite{chen2024disentanglement}.
In addition, the decoupling operation that TAP conducts (specifics outlined in Sec. \ref{sec:tap-method}) differs fundamentally from prior approaches in both selection criterion and problem setting. Existing methods such as \cite{tamirisa2024fedselect} and layer-wise selective \cite{sun2025exploring} fine-tuning operate in task-modality homogeneous settings and rely on parameter or layer-level contribution heuristics to guide replacement. In contrast with existing works, TAP performs decoupling at the modality-task pair level, \textit{unique to this setting}, and leverages a loss-based criterion. Therefore, these methodologies cannot be naively extended to the modality-task foundation model fine-tuning setup. 

\textbf{Parameter Efficient Fine-tuning (PEFT): }In fine-tuning of foundation models, considerable attention has been given to methodologies that efficiently update only a small number of parameters. These techniques fall into the realm of PEFT, and are particularly well-suited for FL scenarios with resource-constrained clients~\citep{wang2019adaptive}. Well-known PEFT methods include prefix tuning~\citep{li2021prefix}, prompt tuning~\citep{lester2021power}, and Low-Rank Adaptation (LoRA)~\citep{hu2021lora}. We show how TAP can invoke PEFT methods for the model fine-tuning piece of foundation model training in multi-modal multi-task FL settings.

\textbf{Knowledge Distillation (KD):} KD is a suite of techniques designed to distill useful knowledge from one model (the teacher) to another (the student). The teacher, which has been trained on data that would be beneficial to one or more tasks the student seeks to optimize, allows the student to utilize its logits to minimize the Kullback-Leibler (KL) divergence~\citep{kullback1951information} between them~\citep{gou2021knowledge}. 
In FL, KD has been shown to be a powerful technique in dealing with the issue non-i.i.d. data distributions between clients~\citep{hsieh2020non, li2020federated, lee2024smart} by engaging with the server's logits on a per-label basis~\citep{jeong2018communication}. Moreover, KD has been shown to act as an effective regularizer in PFL, where the global model's logits are employed to reduce overfitting~\citep{chen2024feddat}. In TAP, we show that KD offers a method by which individual tasks on a client can be augmented with performance gains while maintaining per-task personalization.


\section{Setup and Convergence Analysis}
\subsection{PFL in Multi-Task, Multi-Modality Setup}
\label{sec:setup}

\begin{figure}[t]
    \centering
    \includegraphics[width=0.92\linewidth]{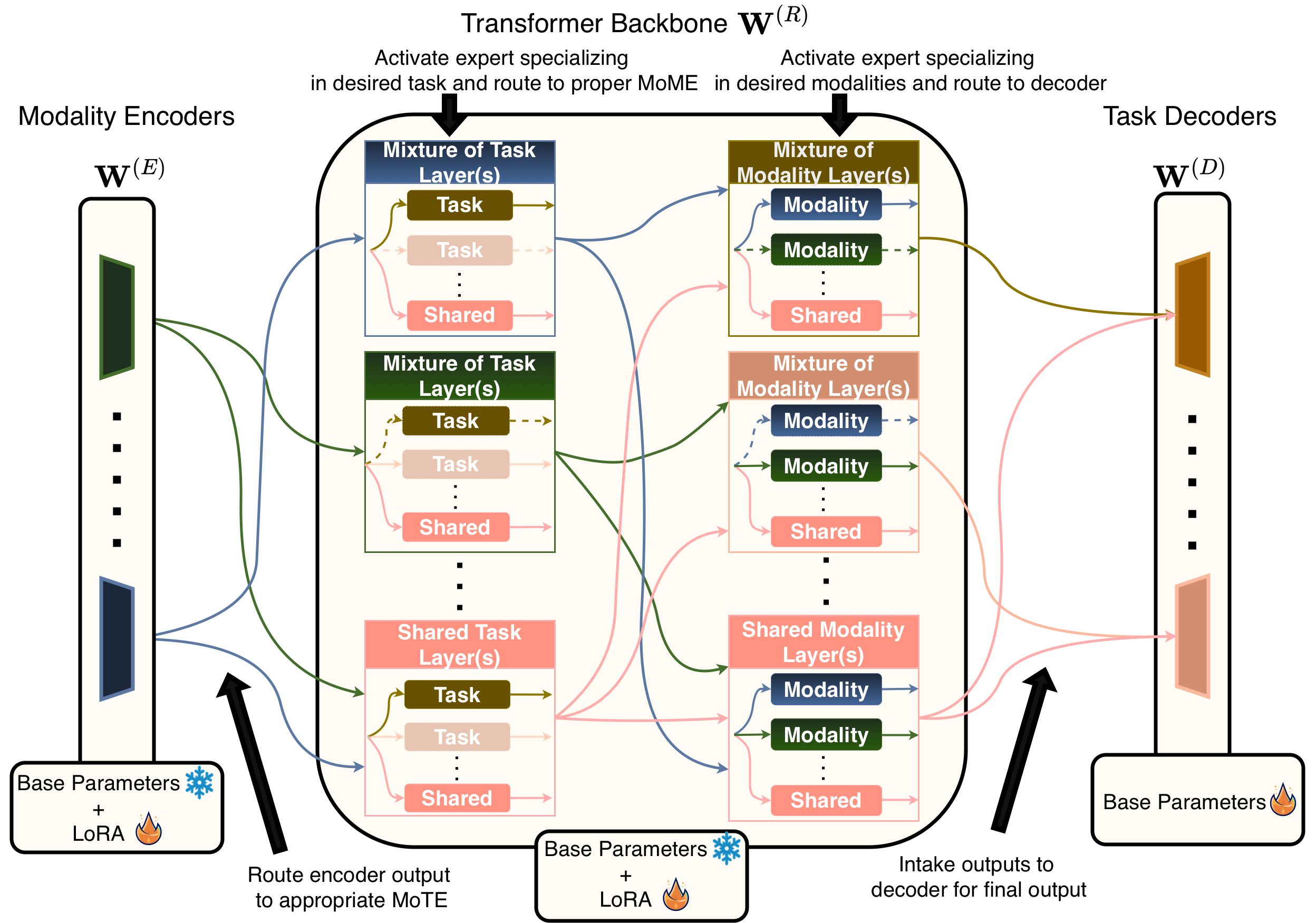}
    \caption{Architecture of server model $\mathbf{W}$. Each local client $c_i$ has a subset of $\mathbf{W}$ consisting of encoders, transformer layers, and decoders relevant to its set of modalities and tasks.}
    \label{fig:model_architecture}
\vspace{-1mm}
\end{figure}

We consider a multi-modal and multi-task PFL setup with $K$ clients in set $c_i \in \mathcal{C}$ and a server $S$, with $\mathcal{C}_{t} \subseteq \mathcal{C}$ clients selected for aggregation each communication round $t$. Adopting the architecture from \cite{chen2024disentanglement}, the server model's parameter vector $\mathbf{W} \in \mathbb{R}^{d \times 1}$ holds all modality encoders $\mathbf{W}^{(E)} \in \mathbb{R}^{d^{(E)} \times 1}$ and task decoders $\mathbf{W}^{(D)} \in \mathbb{R}^{d^{(D)} \times 1}$. Therefore, the server model is designed to handle all modalities $\mathcal{M}$ and tasks $\mathcal{O}$, i.e., $\mathcal{M} = \bigcup_{c_i \in \mathcal{C}} \mathcal{M}_i$ and $\mathcal{O} = \bigcup_{c_i \in \mathcal{C}} \mathcal{O}_i$, where $\mathcal{M}_i$ and $\mathcal{O}_i$ are the modalities and tasks pertaining to client $c_i$ respectively. The transformer backbone $\mathbf{W}^{(R)} \in \mathbb{R}^{d^{(R)} \times 1}$ sits between the encoders and decoders. Additional details on the server model layout can be found in Appendix~\ref{appendix:model-specifics}.

Besides the general architecture, the transformer backbone and encoder parameters are pre-trained and frozen, with these components being fine-tuned with LoRA low-rank matrices $\mathbf{A}$ and $\mathbf{B}$. This means only the decoders are fully trained and tailored to the downstream tasks. Each client $c_i$ holds a subset of the server model, consisting of modality encoders $\mathbf{W}_{[i]}^{(E)} \in \mathbb{R}^{d_i^{(E)} \times 1}$, task decoders $\mathbf{W}_{[i]}^{(D)} \in \mathbb{R}^{d_i^{(D)} \times 1}$, transformer backbone $\mathbf{W}_{[i]}^{(R)} \in \mathbb{R}^{d_i^{(R)} \times 1}$, and LoRA matrices $\mathbf{A}_i$ and $\mathbf{B}_i$. For readability, additional details are moved and provided in Appendix \ref{appendix:model-specifics} (e.g., additional notation definitions).
Fig. \ref{fig:model_architecture} presents the multi-modal, multi-task server model architecture along with the depiction of the LoRA fine-tuned and fully trained components. We note that Fig. \ref{fig:model_architecture} outlines a Mixture of Experts (MoE) \cite{yuksel2012twenty} structure, \textbf{\textit{which is an artifact from the architecture adopted in \cite{chen2024disentanglement} and is not a core contribution of TAP's methodology.}}

For training, clients will train their local models for $\tau$ minibatch iterations, and then broadcast their parameters to the server for aggregation. When the server transmits parameters back to each client after aggregation, it will only send the decoders and LoRA parameters that are relevant to each client based off their tasks and modalities. Utilizing this structure, in traditional federated learning, the objective is find parameters that minimize the global objective, which can be expressed as
\begin{equation}
    \min_{\widetilde{\mathbf{W}}} \left[ f\left( \widetilde{\mathbf{W}} \right) := \frac{1}{|\mathcal{D}|} \sum_{i=1}^{K} |\mathcal{D}_i| \cdot g_i\left(\widetilde{\mathbf{W}}\right) \right],
\label{eq:reg-FL}
\end{equation}
where $\mathcal{D}$ and $\mathcal{D}_i \subseteq \mathcal{D}$ denote the full dataset across all clients and a local dataset on client $c_i$ respectively. We say that the server model trainable parameter vector $\widetilde{\mathbf{W}}$ can be partitioned into disjoint components via $\mathcal{B} = \{\mathcal{B}_0, \ldots, \mathcal{B}_R\}$, which we call ``blocks." Then, for notation purposes, there are projection operators $P_{\mathcal{B}_r}$ (linear) that zero out coordinates outside of block $\mathcal{B}_r$, so $\widetilde{\mathbf{W}}_{[r]} = P_{\mathcal{B}_r} \widetilde{\mathbf{W}}$ and $\widetilde{\mathbf{W}} = \sum_{r=0}^{R} \widetilde{\mathbf{W}}_{[r]}$. Each client $c_i$ owns a subset of blocks $\mathcal{B}_i \subseteq \mathcal{B}$, and since the client's local loss only depends on these blocks, local loss for client $c_i$ is $g_i \left( \widetilde{\mathbf{W}} \right) = f_i \left( P_{\mathcal{B}_i} \widetilde{\mathbf{W}} \right) = f_i \left( \widetilde{\mathbf{W}}_{[i]} \right)$.
Outside of these considerations, when dealing with multi-task scenarios, it is also necessary to consider the loss of multiple differing tasks, resulting in
\begin{equation}
    g_i\left(\widetilde{\mathbf{W}}\right) = \mathbb{E}_{\mathcal{H}_i \sim \mathcal{D}_i} \left[ \sum_{o \in \mathcal{O}_i} \lambda_o \cdot  \ell_{i,o} \left( \mathcal{H}_{i,o} ; \widetilde{\mathbf{W}}_{[i,o]} \right) \right],
\end{equation}
where $\ell_{i,o}$, $\mathcal{H}_{i,o} \subseteq \mathcal{H}_i$, and $\lambda_o$ denotes the sample loss of $c_i$ on task $o$, a subset of minibatch $\mathcal{H}_i$ with samples pertaining to task $o$, and the weighting given to the loss of task $o$ respectively. $\widetilde{\mathbf{W}}_{[i,o]} = P_{\mathcal{B}_{i,o}}\widetilde{\mathbf{W}}$, with $\mathcal{B}_{i, o} \subseteq \mathcal{B}_i$ being a subset of $\mathcal{B}_i$ pertaining to parts relevant to task $o$. However, unlike \eqref{eq:reg-FL}, in PFL, the endpoint is for each client to achieve low local loss, i.e., a low value on $g_i(\cdot)$. Due to heterogeneous model architectures amongst clients, aggregation is performed by utilizing FedAvg~\citep{mcmahan2017communication} on each component individually, combining common components based off the clients chosen for a specific aggregation round. Therefore, it is a possibility that the entirety of $\widetilde{\mathbf{W}}$ is not updated within a single communication round.

\subsection{Convergence Analysis of Server Model}
\label{sec:conv}
In this section, we analyze the convergence of the server model under its modality-task pair architecture. In particular, the following assumptions that are commonly used in the analysis of existing FL literature~\citep{bottou2018optimization, li2020federated, fang2022communication, lee2025cooperative} are made:

\begin{assumption}\label{assump:l-smooth}
     $g_i(\widetilde{\mathbf{W}})$ is differentiable and $L$-smooth, i.e., there exists a positive constant $L$ such that
     $\|\nabla g_i ( \widetilde{\mathbf{W}}_1) - \nabla g_i ( \widetilde{\mathbf{W}}_2)\| \leq  L\|  \widetilde{\mathbf{W}}_1 -  \widetilde{\mathbf{W}}_2\|$
     for any $\widetilde{\mathbf{W}}_1$ and $\widetilde{\mathbf{W}}_2$.
\end{assumption}

\begin{assumption}\label{assump:bounded-variance}
    The minibatch $\mathcal{H}_i$ gradient for client $c_i$, denoted as $\nabla \widetilde{\ell}_i\left( \mathcal{H}_i; \widetilde{\mathbf{W}}_{[i]} \right) = \sum_{o \in \mathcal{O}_i} \lambda_o \cdot \ell_{i,o} \left( \mathcal{H}_{i,o} ; \widetilde{\mathbf{W}}_{[i,o]} \right)$ is an unbiased estimate of $\nabla_{}g_i(\widetilde{\mathbf{W}})$
    and its variance is bounded, i.e.,
    $
    \mathbb{E} \left| \left|\widetilde{\ell}_i\left( \mathcal{H}_i; \widetilde{\mathbf{W}}_{[i]} \right) - \nabla_{}g_i(\widetilde{\mathbf{W}}) \right| \right|^2 \leq \sigma^2.
    $
\end{assumption}

\begin{assumption}\label{assump:bounded-heterogeneity}
    For each block $\mathcal{B}_r$, the block gradient heterogeneity is bounded, i.e.,
    $
    \frac{1}{K_r} \sum_{i: \mathcal{B}_r \in \mathcal{B}_i} \| \nabla g_i(\widetilde{\mathbf{W}}_{[r]}) - \nabla f (\widetilde{\mathbf{W}}_{[r]}) \|^2 \leq \zeta_r^2,
    $
   where $K_r$ is the number of clients having block $\mathcal{B}_r$. 
\end{assumption}

\begin{theorem}\label{theorem:server-bound}
Suppose Assumptions \ref{assump:l-smooth}-\ref{assump:bounded-heterogeneity} hold. Then the iterates generated by server model in Fig. \ref{fig:model_architecture} with full client participation and learning rate $\eta_t \leq \min \left\{ \frac{1}{48L\tau}, \frac{1}{\sqrt{8}L\tau}, \left( \frac{1}{96L^3\tau^3} \right)^{\frac{1}{3}} \right\}$ satisfies:
\begin{align}\label{eq:server-bound}
    \frac{1}{\sum_{t=0}^{T-1} \eta_t } \sum_{t=0}^{T-1} \!\eta_t \mathbb{E} \| \nabla f(\widetilde{\mathbf{W}}_t) \|^2 \leq & \frac{8\! \left( \!f(\widetilde{\mathbf{W}}_0) \!-\! \mathbb{E}\big[f(\widetilde{\mathbf{W}}_T)\big] \!\right)}{\sum_{t=0}^{T-1} \eta_t}
    + \frac{16 L \tau (C_{K} \sigma^2 +3 \tau Z)}{\sum_{t=0}^{T-1} \eta_t}\sum_{t=0}^{T-1} \eta_t^2  \nonumber \\
    &+ \frac{16\tau^3L^2\left(R\sigma^2 +Z \right)}{\sum_{t=0}^{T-1} \eta_t}\sum_{t=0}^{T-1} \eta^3_t,
\end{align}

\end{theorem}
\noindent
where $\widetilde{\mathbf{W}}_0$ and $\widetilde{\mathbf{W}}_T$ are the global parameters at global timestep 0 and $T$ respectively; $\tau$ is the number of local iterations, $R$ is the number of disjoint blocks, $Z := \sum_{r=0}^{R}\zeta_r^2$, and $C_K := \sum_{r=0}^{R} \frac{1}{K_r}$. Note that with a diminishing step size of $\eta_t = \frac{\alpha}{t+1}$, with $\alpha = \min \left\{ \frac{1}{48L\tau}, \frac{1}{\sqrt{8}L\tau}, \left( \frac{1}{96L^3\tau^3} \right)^{\frac{1}{3}} \right\}$, $ \lim_{T\rightarrow \infty} \frac{1}{\sum_{t=0}^{T-1} \eta_t} \rightarrow 0$, $\lim_{T\rightarrow \infty} \frac{1}{\sum_{t=0}^{T-1} \eta_t}\sum_{t=0}^{T-1} \eta_t^2 \rightarrow 0$, and $\lim_{T\rightarrow \infty} \frac{1}{\sum_{t=0}^{T-1} \eta_t}\sum_{t=0}^{T-1} \eta_t^3 \rightarrow 0$. Hence, the RHS of \eqref{eq:server-bound} goes to 0 as $T$ increases to infinity. If there are more disjoint blocks in $\mathcal{B}$ to consider, and therefore more modality-task pairs, then we note that the bound increases in the last two terms of the RHS. This means that if the number of modality-task pairs increases, the ability of the server to serve all clients naturally decreases, which directly motivates TAP's two-step process, which \textbf{\textit{seeks to decouple itself from the FL process unless beneficial}} (presented in Sec. \ref{sec:TAP-methodology}). Step-by-step derivation of the theorem can be found in Appendix \ref{proof:server-bound}.

\section{TAP: Two-Stage Adaptive Personalization}\label{sec:TAP-methodology}
\subsection{Adaptive Replacement for Personalization}
\label{sec:tap-method}

\begin{figure*}[t]
    \centering
    \includegraphics[width=1.0\linewidth]{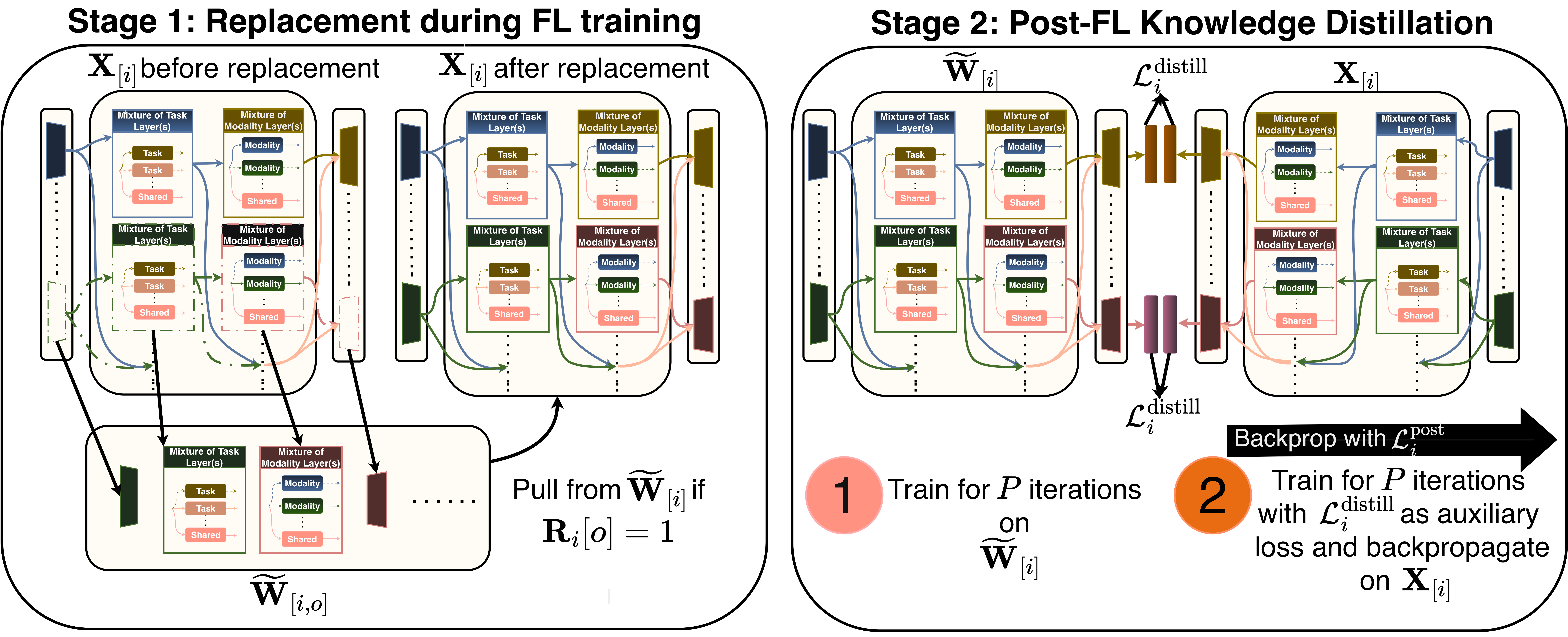}
    \caption{The two-step process of the proposed TAP algorithm. Firstly, during FL training, at each time step, $\mathbf{X}_{[i]}$ will load in parameters from $\widetilde{\mathbf{W}}_{[i, o]}$ when $\mathbf{R}_i [o] = 1$. After FL, using $\widetilde{\mathbf{W}}_{[i]}$ as a teacher, $\mathbf{X}_{[i]}$ will engage in knowledge distillation (KD).}
    \label{fig:two-stage_method}
\end{figure*}
Since the goal of each client $c_i$ is to maximize personalization of its multi-modal and multi-task model towards its own local dataset, relying solely on server parameters $\widetilde{\mathbf{W}}$ is challenging due to fact that $\widetilde{\mathbf{W}}$ is seeking to optimize all tasks $\mathcal{O}$, which utilizes all modalities $\mathcal{M}$. Therefore, it is desirable to limit interaction with $\widetilde{\mathbf{W}}_{[i]}$ (a subset of $\widetilde{\mathbf{W}}$ for client $c_i$), as optimizing for tasks not in $\mathcal{O}_i$ could interfere with personalization, 
\textbf{\textit{which is directly motivated from Theorem \ref{theorem:server-bound}.}} For this reason, on top of $\widetilde{\mathbf{W}}_{[i]}$, which engages in the vanilla FL training protocol, we utilize local personalized parameters $\mathbf{X}_{[i]}$, which follows the same architecture as $\widetilde{\mathbf{W}}_{[i]}$, but does not participate in the transmitting and receiving of parameters from the server, and trains only on the local dataset $\mathcal{D}_i$. However, it could still be the case that the returned server parameters has useful information that could benefit the performance of $\mathbf{X}_{[i]}$, especially early on in the training process, when the most beneficial aspects of a task are learned~\citep{frankle2020early}. Therefore, each client will keep track of two types of values: the average local training loss of $\widetilde{\mathbf{W}}_{[i]}$ and $\mathbf{X}_{[i]}$ for each task that was seen over the course of local training for that round $t$ ($\mathcal{O}_{t,i}^{\text{seen}} \subseteq \mathcal{O}_i$, as samples from all tasks might not be present in a subset of data chosen for local training for an iteration) which are defined as $\ell_{i, o}^{(l)}$ and $\ell_{i, o}^{(p)}$ respectively, where $o$ is a singular task. These values are used to update client $c_i$'s history values, defined as $h_{i, o}^{(l)}$ and $h_{i, o}^{(p)}$ via $h_{i, o}^{(l)} = \ell_{i, o}^{(l)}$ and $h_{i, o}^{(p)} = \ell_{i, o}^{(p)}$ (specifics can be found in the pseudocode at Appendix \ref{appendix:pseudocode}). Then, client $c_i$ will introduce margin hyperparameters $m_{i,o}$ for each task in $\mathcal{O}_i$, and will set a value to an entry of indicator vector $\mathbf{R}_i \in \mathbb{R}^{1 \times |\mathcal{O}_i|}$ via
\begin{equation}
    \mathbf{R}_i [o] =
    \begin{cases}
    1 & \texttt{if } h_{i,o}^{(l)} + m_{i,o} < h_{i,o}^{(p)} \\
    0 & \texttt{otherwise}.
\end{cases}
\label{eq:replacement}
\end{equation}
Intuitively, $\mathbf{R}_i [o]$ is an indicator that $\widetilde{\mathbf{W}}_{[i]}$ at time $t$ achieves superior performance compared to $\mathbf{X}_{[i]}$ on task $o$ by at least $m_{i,o}$, preventing replacement due to noise. In addition, through \eqref{eq:replacement}, we ensure $\mathbf{X}_{[i]}$'s loss is no worse than $\widetilde{\mathbf{W}}_{[i]}$'s on any task $o$, preventing replacements on tasks that do not require it. For these reasons, when $\mathbf{R}_i [o] = 1$, client $c_i$ will replace parameters of $\mathbf{X}_{[i]}$ responsible for task $o$ with parameters from $\widetilde{\mathbf{W}}_{[i]}$, conducting the replacement operation (Stage 1 of Fig. \ref{fig:two-stage_method}) via
\begin{equation}
    \mathbf{X}_{[i,o]} = (1 - \mathbf{R}_i[o]) \mathbf{X}_{[i,o]}
  + \mathbf{R}_i[o] \widetilde{\mathbf{W}}_{[i,o]},
\label{eq:replacement-final}
\end{equation}
where $\mathbf{X}_{[i,o]} \subseteq \mathbf{X}_{[i]}$ is the subset of $\mathbf{X}_{[i]}$ responsible for task $o$. In this way, the replacement operation can selectively extract the desired parameters while minimizing interference from other modality-task pairings, thereby enabling effective multi-modal and multi-task personalization.

\textbf{Remark:} We note that the replacement process, which occurs in conjunction with the FL training protocol, does not induce additional communication costs, as the personalized parameters $\mathbf{X}_{[i]}$ \textit{are not transmitted to server $S$}. Moreover, $\mathbf{X}_{[i]}$ can be trained on $c_i$ \textit{in parallel while $\widetilde{\mathbf{W}}_{[i]}$ is being aggregated at $S$}, enabling effective usage of both local and server resources simultaneously.

\subsection{Post-FL Knowledge Distillation}
\label{sec:tap-KD-step}
After FL training has been completed, TAP entails a knowledge distillation (KD)~\citep{gou2021knowledge} phase, whereby a teacher model seeks to distill knowledge to a student model via an auxiliary KL-based~\citep{kullback1951information} loss. Unlike existing works \cite{chen2024feddat}, KD takes place \textit{after FL} as in a modality-task heterogeneous specific setup, \textit{transfer of shared knowledge must happen in a manner that does not reintroduce cross-task interference.} The main reason for the post-FL KD is that, after FL, the global model has stabilized, making it the most appropriate time to engage in KD and is motivated directly by the modality-task heterogeneous setting. In a homogeneous case, such considerations are not needed, as \textit{cross-modality, cross-task interference is not a consideration.} As a first step of the post-FL KD process, the final parameters $\widetilde{\mathbf{W}}_{[i]}$ that were returned from the server $S$ will be trained locally for $P$ mini-batch iterations, and will then be used as a teacher model for $\mathbf{X}_{[i]}$, who will then also train for $P$ iterations utilizing KD. The intuition is that $\widetilde{\mathbf{W}}_{[i]}$ now benefits from both the FL process (which captures generalized knowledge) and some specialization on local data, free from conflicts introduced by other clients' distributions. 
Unlike more aggressive operations such as replacement, post-FL KD enables transfer of beneficial knowledge while preserving each client’s personalized nature. 
We define the distillation loss for each $c_i$ as 
\begin{align}
\mathcal{L}_i^{\text{distill}}(\mathcal{H}_{i,o})
= \widetilde{\tau}^2 \cdot
\mathrm{KL}\!\left(
\operatorname{softmax}\!\left(\tfrac{f_{\text{dec}}^{(\tilde{t})}(f_{\text{bb}}^{(\tilde{t})}(f_\text{enc}^{(\tilde{t})}(\mathcal{H}_{i,o})))}{\widetilde{\tau}}\right)
\;\middle\|\;
\operatorname{softmax}\!\left(\tfrac{f_{\text{dec}}^{(\tilde{s})}(f_{\text{bb}}^{(\tilde{s})}(f_\text{enc}^{(\tilde{s})}(\mathcal{H}_{i,o})))}{\widetilde{\tau}}\right)
\right),
\label{eq:kd-eq}
\end{align}
where $f_{\text{enc}}$, $f_{\text{bb}}$, and $f_{\text{dec}}$ are outputs from the encoders, backbone, and decoders, with $(\tilde{t})$ and $(\tilde{s})$ denoting the teacher (global) and student (personalized) respectively; and $\widetilde{\tau}$ is temperature coefficient. During implementation, we make the student log-softmax, which is common practice in popular frameworks (e.g., PyTorch).
Utilizing \eqref{eq:kd-eq}, $\mathbf{X}_{[i]}$ is trained in the post-FL phase (Stage 2 of Fig. \ref{fig:two-stage_method}) via
\begin{align}
\mathcal{L}_i^{\text{post}} \left( \mathcal{H}_i \right)
= \sum_{o \in \mathcal{O}'_i} \lambda_o \cdot \,\ell_{i,o}(\mathcal{H}_{i,o})
+ \sum_{o \in \mathcal{O}'_i} \beta_{o} \cdot\, \mathcal{L}_i^{\text{distill}} \left(\mathcal{H}_{i, o}\right),
\label{eq:kd-loss}
\end{align}
where $\beta_o$ and $\mathcal{O}'_i \subseteq \mathcal{O}_i$ is the weight given to the KD loss for task $o$ and the set of tasks present in minibatch $\mathcal{H}_i$ respectively. Pseudocode of the TAP algorithm can be found in Appendix~\ref{appendix:pseudocode}.

\section{Experimental Results}\label{sec:experiments}

\subsection{Setup}
\label{exp:setup}

\textbf{Model Architecture:} To evaluate TAP, we consider a server model with text and image input modalities, with tasks relating to image generation, text generation, image classification, and text classification. We consider two pre-trained foundation models dealing with image and text modalities: FLAVA~\citep{singh2022flava} and ViLT~\citep{kim2021vilt}. For FLAVA, the modality encoders are the pre-trained image and text encoders, with ViLT's composing of its linear patch projections and word embeddings. In terms of the transformer backbone, due to existing work suggesting that the pruning of later layers of a pre-trained foundation model~\citep{sajjad2023effect} can be conducted while maintaining a majority of its performance, we load the first two layers of the multi-modal encoder of FLAVA into $\mathbf{W}^{(R)}$, with same number applied to ViLT. Moreover, within the backbone, similar to existing work~\citep{wu2022residual}, we load the pre-trained feed forward network (FFN) weights at each layer as the frozen pre-trained weights of the MoE. As outlined in Sec. \ref{sec:setup}, the encoders and transformer backbone are fine-tuned with LoRA, with specifics outlined in Appendix \ref{appendix:hyperparameters}. For the decoders, the generation heads are shared for all generation tasks of a particular modality (decoder specifics outlined in Appendix \ref{appendix:decoder-arch}). In all experiments, the AdamW optimizer is used~\citep{loshchilov2017decoupled}. More specifics can be found in Appendix \ref{appendix:implementation-specs}.

\textbf{Datasets:} We use eight common datasets for FLAVA and six for ViLT, spread across 30 clients. We utilize Tiny-Imagenet~\citep{le2015tiny} and CIFAR-100~\citep{Krizhevsky09learningmultiple} for image classification, Fashion-MNIST (FMNIST)~\citep{xiao2017fashion} and Caltech-256~\citep{griffin2007caltech} for image generation reconstruction, AG News for text classification~\citep{zhang2015character, ag_news_corpus}, and Massive Multitask Language Understanding (MMLU)~\citep{hendrycks2021ethics, hendryckstest2021} (Professional Law, Professional Medicine, and Moral Scenario subsets), the small variant of VQAv2 (Visual Question Answering v2)~\citep{goyal2017making}, and CommonGen~\citep{lin-etal-2020-commongen} for text generation. For measuring the quality of the generated text, we use commonly utilized metrics of BERTScore (BS)~\citep{zhang2019bertscore} and METEOR~\citep{banarjee2005}. While FLAVA uses all 8 datasets, we exclude the usage of Tiny-ImageNet and CIFAR-100 for ViLT, as ViLT's pre-trained model is not well-suited for such tasks.

\textbf{Baselines:} To benchmark the performance of TAP, we compare against the following baselines: Local, FedAvg~\citep{mcmahan2017communication}, Per-FedAvg~\citep{fallah2020personalized}, DisentAFL~\citep{chen2024disentanglement}, and FedDAT~\citep{chen2024feddat}. For local, no FL takes place, with each client training its own model locally without any interaction from the server. With FedAvg, as mentioned in Sec. \ref{sec:setup}, server $S$ aggregates common parameter components from the subset of clients chosen at round $t$ and returns the parameters relevant to each client's modality and tasks back down to all clients $c_i \in \mathcal{C}$. Per-FedAvg engages in an inner and outer step for updating the model, to find an initial shared point that can help clients adapt quickly to their own data. DisentAFL, besides utilizing the model architectures of server $S$ and clients $c_i \in \mathcal{C}$ as outlined in Fig. \ref{fig:model_architecture} and Sec. \ref{sec:setup}, also introduces a disentanglement loss to local training, whereby it attempts to make differing modality-task pairs in the latent space orthogonal. FedDAT introduces adapters near the FNNs of a pre-trained model, whose structure includes a global and personalized component. Lastly, we also incorporate post-FL baselines for FedAvg, DisentAFL, Per-FedAvg, and FedDAT, where $\widetilde{\mathbf{W}}_{[i]}$ is trained for $P$ iterations on local dataset $\mathcal{D}_i$ after FL. This is to ascertain whether TAP's two-stage process, which includes a KD-based post-FL phase, is superior to extending an existing PFL method by training for $P$ iterations on local data.

\textbf{Hardware and Hyperparameters:} For all results, experiments were conducted on a server with a cluster of four NVIDIA A100-40GB GPUs. For the weighting of each task's loss during a minibatch iteration, we tune each task's weight $\lambda_o$ based on the number of samples related to task $o$ within a batch, i.e., $\frac{|\mathcal{H}_{i,o}|}{|\mathcal{H}_i|}$. For other hyperparameters, explicit details are outlined in Appendix \ref{appendix:hyperparameters}.

\textbf{Additional Experimental Results:} Additional experimental results (e.g., ablation studies, number of replacements on differing margins, etc.) can be found in Appendix \ref{appendix:additional-exp}.

\subsection{Results}
\label{exp:main-results}

\begin{table*}[t]
\centering
\caption{Performance comparison across FLAVA and ViLT on image datasets. N/A columns denote no experiments being conducted due to the pre-trained model not being well-suited for certain tasks.}
\resizebox{\linewidth}{!}{%
\label{table:main-img-combined}
\begin{tabular}{l l c c c c c c}
\toprule
\textbf{Model} & \textbf{Method} 
& \textbf{Tiny-ImageNet} 
& \textbf{CIFAR-100} 
& \textbf{FMNIST} 
& \textbf{Caltech-256} 
& \textbf{Avg. Class.} 
& \textbf{Avg. Gen.} \\
\cmidrule(lr){3-8}
& 
& \textbf{Acc} $(\uparrow)$ 
& \textbf{Acc} $(\uparrow)$ 
& \textbf{MSE} $(\downarrow)$ 
& \textbf{MSE} $(\downarrow)$ 
& \textbf{Acc} $(\uparrow)$ 
& \textbf{MSE} $(\downarrow)$ \\
\midrule

\multirow{10}{*}{\textbf{FLAVA}}
& Local & $25.35 \scriptstyle\pm 12.23$ & $32.01 \scriptstyle\pm 11.02$ & $0.6040 \scriptstyle\pm 0.0179$ & $0.6479 \scriptstyle\pm 0.0842$ & $28.68$ & $0.6259$ \\
& FedAvg~\citep{mcmahan2017communication} & $30.66 \scriptstyle\pm 6.59$ & $46.20 \scriptstyle\pm 7.45$ & $0.6109 \scriptstyle\pm 0.0261$ & $0.4595 \scriptstyle\pm 0.0970$ & $38.43$ & $0.5352$ \\
& FedAvg + Post-train & $41.51 \scriptstyle\pm 7.28$ & $\underline{54.22} \scriptstyle\pm 5.34$ & $0.5553 \scriptstyle\pm 0.0071$ & $0.4005 \scriptstyle\pm 0.0326$ & $\underline{47.86}$ & $0.4779$ \\
& Per-FedAvg~\citep{fallah2020personalized} & $24.97 \scriptstyle\pm 8.41$ & $40.53 \scriptstyle\pm 10.04$ & $0.7717 \scriptstyle\pm 0.0981$ & $0.6379 \scriptstyle\pm 0.0795$ & $32.75$ & $0.7048$ \\
& Per-FedAvg + Post-train & $40.60 \scriptstyle\pm 5.69$ & $53.15 \scriptstyle\pm 5.10$ & $0.5805 \scriptstyle\pm 0.0079$ & $0.4034 \scriptstyle\pm 0.0303$ & $46.87$ & $0.4919$ \\
& DisentAFL~\citep{chen2024disentanglement} & $38.29 \scriptstyle\pm 10.03$ & $1.07 \scriptstyle\pm 0.23$ & $0.9085 \scriptstyle\pm 0.0177$ & $0.5526 \scriptstyle\pm 0.0411$ & $19.68$ & $0.7305$ \\
& DisentAFL + Post-train & $\underline{46.62} \scriptstyle\pm 7.82$ & $4.33 \scriptstyle\pm 0.65$ & $0.7369 \scriptstyle\pm 0.0390$ & $0.5265 \scriptstyle\pm 0.0385$ & $25.48$ & $0.6316$ \\
& FedDAT~\citep{chen2024feddat} & $15.43 \scriptstyle\pm 6.49$ & $30.23 \scriptstyle\pm 10.74$ & $0.6031 \scriptstyle\pm 0.0167$ & $0.4187 \scriptstyle\pm 0.0696$ & $22.83$ & $0.5109$ \\
& FedDAT + Post-train & $25.22 \scriptstyle\pm 13.68$ & $39.22 \scriptstyle\pm 13.75$ & $\underline{0.5506} \scriptstyle\pm 0.0041$ & $\textbf{0.3715} \scriptstyle\pm 0.0106$ & $32.22$ & $\textbf{0.4610}$ \\
\rowcolor{blue!10}
& \textbf{TAP (Ours)} & $\textbf{47.30} \scriptstyle\pm 6.57$ & $\textbf{56.84} \scriptstyle\pm 4.70$ & $\textbf{0.5478} \scriptstyle\pm 0.0074$ & $\underline{0.3899} \scriptstyle\pm 0.0305$ & $\textbf{52.07}$ & $\underline{0.4688}$ \\

\midrule

\multirow{10}{*}{\textbf{ViLT}}
& Local & N/A & N/A & $2.9185 \scriptstyle\pm 1.8431$ & $1.7164 \scriptstyle\pm 1.5459$ & N/A & $2.3174$ \\
& FedAvg~\citep{mcmahan2017communication} & N/A & N/A & $0.6330 \scriptstyle\pm 0.0412$ & $0.5784 \scriptstyle\pm 0.0521$ & N/A & $0.6057$ \\
& FedAvg + Post-train & N/A & N/A & $0.6067 \scriptstyle\pm 0.0187$ & $0.5354 \scriptstyle\pm 0.0301$ & N/A & $0.5710$ \\
& Per-FedAvg~\citep{fallah2020personalized} & N/A & N/A & $0.6963 \scriptstyle\pm 0.0602$ & $0.7012 \scriptstyle\pm 0.0378$ & N/A & $0.6987$ \\
& Per-FedAvg + Post-train & N/A & N/A & $0.6143 \scriptstyle\pm 0.0229$ & $0.5500 \scriptstyle\pm 0.0230$ & N/A & $0.5821$ \\
& DisentAFL~\citep{chen2024disentanglement} & N/A & N/A & $4.5943 \scriptstyle\pm 2.7091$ & $3.1483 \scriptstyle\pm 2.4651$ & N/A & $3.8712$ \\
& DisentAFL + Post-train & N/A & N/A & $3.0998 \scriptstyle\pm 1.6832$ & $2.8579 \scriptstyle\pm 2.2146$ & N/A & $2.9788$ \\
& FedDAT~\citep{chen2024feddat} & N/A & N/A & $0.6152 \scriptstyle\pm 0.0313$ & $0.5638 \scriptstyle\pm 0.0469$ & N/A & $0.5894$ \\
& FedDAT + Post-train & N/A & N/A & $\textbf{0.5976} \scriptstyle\pm 0.0221$ & $\textbf{0.5140} \scriptstyle\pm 0.0141$ & N/A & $\textbf{0.5558}$ \\
\rowcolor{blue!10}
& \textbf{TAP (Ours)} & N/A & N/A & $\underline{0.5992} \scriptstyle\pm 0.0117$ & $\underline{0.5300} \scriptstyle\pm 0.0247$ & N/A & $\underline{0.5646}$ \\

\bottomrule
\end{tabular}%
}
\end{table*}

\begin{table*}[t] 
\centering
\caption{Performance comparison across FLAVA and ViLT on text datasets.}
\resizebox{\linewidth}{!}{%
\label{table:main-txt}
\begin{tabular}{l l c cc cc cc cc}
\toprule
\textbf{Model} & \textbf{Method} & \textbf{AG News} & \multicolumn{2}{c}{\textbf{MMLU}} & \multicolumn{2}{c}{\textbf{VQA}} & \multicolumn{2}{c}{\textbf{CommonGen}} & \multicolumn{2}{c}{\textbf{Avg. Gen.}} \\
\cmidrule(lr){3-11}
& & \textbf{Acc} $(\uparrow)$ & \textbf{BS} $(\uparrow)$ & \textbf{METEOR} $(\uparrow)$ & \textbf{BS} $(\uparrow)$ & \textbf{METEOR} $(\uparrow)$ & \textbf{BS} $(\uparrow)$ & \textbf{METEOR} $(\uparrow)$ & \textbf{BS} $(\uparrow)$ & \textbf{METEOR} $(\uparrow)$ \\
\midrule

\multirow{6}{*}{\centering\textbf{FLAVA}}
& Local & $90.25 \scriptstyle\pm 0.76$ & $41.63 \scriptstyle\pm 1.26$ & $\underline{20.89} \scriptstyle\pm 2.83$ & $47.72 \scriptstyle\pm 1.69$ & $10.44 \scriptstyle\pm 2.37$ & $30.69 \scriptstyle\pm 1.37$ & $13.20 \scriptstyle\pm 2.71$ & $39.69$ & $13.64$ \\
& FedAvg~\citep{mcmahan2017communication} & $93.02 \scriptstyle\pm 0.13$ & $37.38 \scriptstyle\pm 3.37$ & $6.22 \scriptstyle\pm 1.52$ & $38.70 \scriptstyle\pm 12.70$ & $12.76 \scriptstyle\pm 10.35$ & $30.29 \scriptstyle\pm 3.67$ & $9.34 \scriptstyle\pm 4.81$ & $35.07$ & $10.08$ \\
& FedAvg + Post-train & $\textbf{93.05} \scriptstyle\pm 0.19$ & $38.34 \scriptstyle\pm 5.07$ & $8.35 \scriptstyle\pm 2.13$ & $45.77 \scriptstyle\pm 9.82$ & $23.73 \scriptstyle\pm 11.25$ & $32.35 \scriptstyle\pm 2.68$ & $10.34 \scriptstyle\pm 3.49$ & $38.92$ & $15.30$ \\
& Per-FedAvg~\citep{fallah2020personalized} & $92.62 \scriptstyle\pm 0.30$ & $39.88 \scriptstyle\pm 1.93$ & $8.91 \scriptstyle\pm 1.41$ & $44.09 \scriptstyle\pm 13.71$ & $17.35 \scriptstyle\pm 13.38$ & $32.15 \scriptstyle\pm 4.28$ & $10.38 \scriptstyle\pm 5.19$ & $38.48$ & $12.88$ \\
& Per-FedAvg + Post-train & $\underline{93.03} \scriptstyle\pm 0.20$ & $39.37 \scriptstyle\pm 2.93$ & $11.05 \scriptstyle\pm 4.81$ & $48.11 \scriptstyle\pm 11.70$ & $23.85 \scriptstyle\pm 12.27$ & $32.73 \scriptstyle\pm 3.22$ & $12.62 \scriptstyle\pm 6.94$ & $40.21$ & $16.80$ \\
& DisentAFL~\citep{chen2024disentanglement} & $92.96 \scriptstyle\pm 0.21$ & $40.10 \scriptstyle\pm 0.77$ & $19.54 \scriptstyle\pm 0.85$ & $30.77 \scriptstyle\pm 0.64$ & $5.46 \scriptstyle\pm 0.11$ & $26.67 \scriptstyle\pm 0.48$ & $7.60 \scriptstyle\pm 0.38$ & $30.99$ & $9.13$ \\
& DisentAFL + Post-train & $92.86 \scriptstyle\pm 0.25$ & $40.29 \scriptstyle\pm 0.29$ & $20.37 \scriptstyle\pm 1.58$ & $34.88 \scriptstyle\pm 0.71$ & $7.71 \scriptstyle\pm 0.87$ & $27.80 \scriptstyle\pm 1.91$ & $8.60 \scriptstyle\pm 0.35$ & $33.13$ & $10.60$ \\
& FedDAT~\citep{chen2024feddat} & $92.60 \scriptstyle\pm 0.48$ & $41.67 \scriptstyle\pm 2.73$ & $7.04 \scriptstyle\pm 1.27$ & $50.16 \scriptstyle\pm 18.20$ & $28.43 \scriptstyle\pm 22.66$ & $\underline{37.12} \scriptstyle\pm 6.06$ & $13.87 \scriptstyle\pm 11.83$ & $43.25$ & $18.33$ \\
& FedDAT + Post-train & $92.80 \scriptstyle\pm 0.40$ & $\underline{41.80} \scriptstyle\pm 2.66$ & $10.81 \scriptstyle\pm 6.43$ & $\underline{53.84} \scriptstyle\pm 15.29$ & $\underline{40.07} \scriptstyle\pm 25.74$ & $\textbf{39.79} \scriptstyle\pm 2.12$ & $\textbf{16.42} \scriptstyle\pm 8.36$ & $\underline{45.81}$ & $\underline{24.76}$ \\
\rowcolor{blue!10}
& \textbf{TAP (Ours)} & $92.83 \scriptstyle\pm 0.26$ & $\textbf{46.75} \scriptstyle\pm 5.58$ & $\textbf{21.93} \scriptstyle\pm 4.77$ & $\textbf{69.62} \scriptstyle\pm 6.83$ & $\textbf{48.74} \scriptstyle\pm 14.46$ & $35.61 \scriptstyle\pm 3.34$ & $\underline{15.12} \scriptstyle\pm 2.60$ & $\textbf{51.44}$ & $\textbf{29.93}$ \\
\midrule

\multirow{6}{*}{\centering\textbf{ViLT}}
& Local & $45.99 \scriptstyle\pm 3.97$ & $54.87 \scriptstyle\pm 1.95$ & $39.46 \scriptstyle\pm 2.07$ & $\underline{58.33} \scriptstyle\pm 1.95$ & $29.81 \scriptstyle\pm 5.78$ & $\underline{41.44} \scriptstyle\pm 1.28$ & $\underline{25.31} \scriptstyle\pm 2.30$ & $\underline{50.88}$ & $\underline{29.94}$ \\
& FedAvg~\citep{mcmahan2017communication} & $56.62 \scriptstyle\pm 4.18$ & $44.00 \scriptstyle\pm 1.61$ & $11.93 \scriptstyle\pm 5.99$ & $52.61 \scriptstyle\pm 15.44$ & $42.85 \scriptstyle\pm 22.14$ & $36.79 \scriptstyle\pm 2.60$ & $11.70 \scriptstyle\pm 8.50$ & $44.56$ & $24.20$ \\
& FedAvg + Post-train & $58.08 \scriptstyle\pm 3.31$ & $46.71 \scriptstyle\pm 4.65$ & $13.50 \scriptstyle\pm 8.49$ & $56.71 \scriptstyle\pm 13.27$ & $42.30 \scriptstyle\pm 21.33$ & $38.78 \scriptstyle\pm 2.29$ & $12.33 \scriptstyle\pm 8.31$ & $47.54$ & $24.55$ \\
& Per-FedAvg~\citep{fallah2020personalized} & $56.50 \scriptstyle\pm 3.05$ & $43.95 \scriptstyle\pm 2.86$ & $12.53 \scriptstyle\pm 8.03$ & $50.41 \scriptstyle\pm 16.76$ & $41.73 \scriptstyle\pm 22.32$ & $35.61 \scriptstyle\pm 2.13$ & $10.15 \scriptstyle\pm 6.05$ & $43.20$ & $23.26$ \\
& Per-FedAvg + Post-train & $60.41 \scriptstyle\pm 2.72$ & $45.79 \scriptstyle\pm 4.70$ & $13.02 \scriptstyle\pm 10.12$ & $54.03 \scriptstyle\pm 14.51$ & $41.45 \scriptstyle\pm 21.30$ & $38.27 \scriptstyle\pm 2.91$ & $12.08 \scriptstyle\pm 8.12$ & $46.08$ & $24.02$ \\
& DisentAFL~\citep{chen2024disentanglement} & $56.25 \scriptstyle\pm 3.95$ & $54.67 \scriptstyle\pm 1.71$ & $39.94 \scriptstyle\pm 2.27$ & $41.94 \scriptstyle\pm 0.94$ & $21.30 \scriptstyle\pm 1.25$ & $32.93 \scriptstyle\pm 0.80$ & $9.62 \scriptstyle\pm 0.21$ & $40.89$ & $20.36$ \\
& DisentAFL + Post-train & $56.94 \scriptstyle\pm 3.44$ & $\underline{55.16} \scriptstyle\pm 0.52$ & $\textbf{40.78} \scriptstyle\pm 0.42$ & $43.71 \scriptstyle\pm 1.78$ & $22.74 \scriptstyle\pm 1.71$ & $33.76 \scriptstyle\pm 1.15$ & $9.96\scriptstyle \pm 0.20$ & $42.02$ & $21.23$ \\
& FedDAT~\citep{chen2024feddat} & $\underline{84.97} \scriptstyle\pm 0.48$ & $45.58 \scriptstyle\pm 4.00$ & $13.37 \scriptstyle\pm 9.25$ & $52.14 \scriptstyle\pm 15.63$ & $\underline{42.98} \scriptstyle\pm 22.55$ & $37.38 \scriptstyle\pm 2.70$ & $10.43 \scriptstyle\pm 8.12$ & $44.93$ & $24.04$ \\
& FedDAT + Post-train & $\textbf{84.99} \scriptstyle\pm 0.43$ & $46.16 \scriptstyle\pm 4.26$ & $13.00 \scriptstyle\pm 8.07$ & $54.96 \scriptstyle\pm 14.10$ & $42.28 \scriptstyle\pm 21.49$ & $38.57 \scriptstyle\pm 2.65$ & $11.99 \scriptstyle\pm 8.29$ & $46.65$ & $24.31$ \\
\rowcolor{blue!10}
& \textbf{TAP (Ours)} & $58.42 \scriptstyle\pm 3.52$ & $\textbf{57.83} \scriptstyle\pm 3.87$ & $\underline{40.74} \scriptstyle\pm 1.04$ & $\textbf{77.80} \scriptstyle\pm 3.95$ & $\textbf{60.53} \scriptstyle\pm 12.60$ & $\textbf{41.93} \scriptstyle\pm 1.05$ & $\textbf{25.54} \scriptstyle\pm 1.46$ & $\textbf{59.46}$ & $\textbf{42.58}$ \\
\bottomrule
\end{tabular}%
}
\end{table*}

\begin{table*}[t]
\centering
\caption{Ablation study on the use of KD in the post-training phase on image datasets. N/A columns denote no experiments being conducted due to the pre-trained model not being well-suited for certain tasks.}
\resizebox{\linewidth}{!}{%
\label{table:kd-combined-img}
\begin{tabular}{l l c c c c c c}
\toprule
\textbf{Model} & \textbf{Method} 
& \textbf{Tiny-ImageNet} 
& \textbf{CIFAR-100} 
& \textbf{FMNIST} 
& \textbf{Caltech-256} 
& \textbf{Avg. Class.} 
& \textbf{Avg. Gen.} \\
\cmidrule(lr){3-8}
& 
& \textbf{Acc} $(\uparrow)$ 
& \textbf{Acc} $(\uparrow)$ 
& \textbf{MSE} $(\downarrow)$ 
& \textbf{MSE} $(\downarrow)$ 
& \textbf{Acc} $(\uparrow)$ 
& \textbf{MSE} $(\downarrow)$ \\
\midrule

\rowcolor{gray!5}
\multirow{1}{*}{\centering\textbf{FLAVA}}
& No KD 
& $\textbf{47.30} \scriptstyle\pm 6.57$ 
& $56.82 \scriptstyle\pm 4.71$ 
& $\textbf{0.5476} \scriptstyle\pm 0.0075$ 
& $\textbf{0.3894} \scriptstyle\pm 0.0307$ 
& $52.06$ 
& $\textbf{0.4685}$ \\

\rowcolor{gray!5}
& KD 
& $\textbf{47.30} \scriptstyle\pm 6.57$ 
& $\textbf{56.84} \scriptstyle\pm 4.70$ 
& $0.5478 \scriptstyle\pm 0.0074$ 
& $0.3899 \scriptstyle\pm 0.0305$ 
& $\textbf{52.07}$ 
& $0.4688$ \\

\midrule

\rowcolor{gray!5}
\multirow{1}{*}{\centering\textbf{ViLT}}
& No KD 
& N/A 
& N/A 
& $\textbf{0.5991} \scriptstyle\pm 0.0117$ 
& $\textbf{0.5300} \scriptstyle\pm 0.0248$ 
& N/A 
& $\textbf{0.5645}$ \\

\rowcolor{gray!5}
& KD 
& N/A 
& N/A 
& $0.5992 \scriptstyle\pm 0.0117$ 
& $\textbf{0.5300} \scriptstyle\pm 0.0247$ 
& N/A 
& $0.5646$ \\

\bottomrule
\end{tabular}%
}
\end{table*}

\begin{table*}[t] 
\centering
\caption{Ablation study on the use of KD in the post-training phase on text datasets.}
\resizebox{\linewidth}{!}{%
\label{table:kd-txt}
\begin{tabular}{l l c cc cc cc cc}
\toprule
\textbf{Model} & \textbf{Method} & \textbf{AG News} & \multicolumn{2}{c}{\textbf{MMLU}} & \multicolumn{2}{c}{\textbf{VQA}} & \multicolumn{2}{c}{\textbf{CommonGen}} & \multicolumn{2}{c}{\textbf{Avg. Gen.}} \\
\cmidrule(lr){3-11}
& & \textbf{Acc} $(\uparrow)$ & \textbf{BS} $(\uparrow)$ & \textbf{METEOR} $(\uparrow)$ & \textbf{BS} $(\uparrow)$ & \textbf{METEOR} $(\uparrow)$ & \textbf{BS} $(\uparrow)$ & \textbf{METEOR} $(\uparrow)$ & \textbf{BS} $(\uparrow)$ & \textbf{METEOR} $(\uparrow)$ \\
\midrule

\rowcolor{gray!5}
\multirow{1}{*}{\centering\textbf{FLAVA}}
& No KD & $92.82 \scriptstyle\pm 0.30$ & $41.53 \scriptstyle\pm 1.26$ & $16.66 \scriptstyle\pm 3.20$ & $51.81 \scriptstyle\pm 4.09$ & $17.38 \scriptstyle\pm 6.16$ & $32.69 \scriptstyle\pm 2.74$ & $14.51 \scriptstyle\pm 2.92$ & $42.11$ & $16.09$ \\
\rowcolor{gray!5}
& KD & $\textbf{92.83} \scriptstyle\pm 0.26$ & $\textbf{46.75} \scriptstyle\pm 5.58$ & $\textbf{21.93} \scriptstyle\pm 4.77$ & $\textbf{69.62} \scriptstyle\pm 6.83$ & $\textbf{48.74} \scriptstyle\pm 14.46$ & $\textbf{35.61} \scriptstyle\pm 3.34$ & $\textbf{15.12} \scriptstyle\pm 2.60$ & $\textbf{51.44}$ & $\textbf{29.93}$ \\
\midrule

\rowcolor{gray!5}
\multirow{1}{*}{\centering\textbf{ViLT}}
& No KD & $\textbf{58.48} \scriptstyle\pm 3.57$ & $54.94 \scriptstyle\pm 1.63$ & $39.94 \scriptstyle\pm 0.89$ & $76.73 \scriptstyle\pm 4.16$ & $58.88 \scriptstyle\pm 12.72$ & $41.59 \scriptstyle\pm 1.23$ & $25.36 \scriptstyle\pm 2.02$ & $58.32$ & $41.68$ \\
\rowcolor{gray!5}
& KD & $58.42 \scriptstyle\pm 3.52$ & $\textbf{57.83} \scriptstyle\pm 3.87$ & $\textbf{40.74} \scriptstyle\pm 1.04$ & $\textbf{77.80} \scriptstyle\pm 3.95$ & $\textbf{60.53} \scriptstyle\pm 12.60$ & $\textbf{41.93} \scriptstyle\pm 1.05$ & $\textbf{25.54} \scriptstyle\pm 1.46$ & $\textbf{59.46}$ & $\textbf{42.58}$ \\
\bottomrule
\end{tabular}%
}
\end{table*}

\textbf{Performances in Comparison to Baselines:} Firstly, we assess the effectiveness of TAP in comparison to the baselines outlined in Sec. \ref{exp:setup}, with results presented in Table \ref{table:main-img-combined} for image-related tasks and Table \ref{table:main-txt} for text-related tasks. Results outlined in bold signify the highest performing method, with the second best underlined. Based off the results from Tables \ref{table:main-img-combined} and \ref{table:main-txt}, we firstly note the superiority of the proposed TAP methodology across a vast majority of the tasks evaluated, with the highest average accuracy and generation scores for image classification and text generation tasks on FLAVA, and with text generation for ViLT. For tasks where performance of TAP lags behind the baselines, TAP still retains high performance and almost always remains close with the best baseline (e.g., Caltech-256 on FLAVA with 0.3715 (FedDAT + Post-train) vs. 0.3899 (TAP), FMNIST on ViLT with 0.5976 (FedDAT + Post-train) vs. 0.5992 (TAP), or MMLU METEOR on ViLT with 40.78 (DisentAFL + Post-train) vs. 40.74 (TAP)), with it often being the next best option available (noted via underline). Moreover, we note that TAP \textit{\textbf{is most consistent algorithm in performance across all tasks,}} whereas some baselines seem to excel at specific tasks while suffering on others (e.g., image generation vs. classification on FLAVA with FedDAT or image generation vs. MMLU on ViLT with DisentAFL). 
An analysis of the statistical significance of the results can be found in Appendix~\ref{appendix:welch-t-test}.


\textbf{Ablation Study on KD:} In this section, we explore the impact of utilizing KD from \eqref{eq:kd-loss} in Tables \ref{table:kd-combined-img} and \ref{table:kd-txt}. We can note from these tables the superiority of utilizing KD in improving the performance, especially on BS and METEOR. For example, the average score increase in BS is 9 to 11, with 1 to 13 for METEOR. For cases where the KD underperforms simple post-training without KD, \textit{the difference is minimal.} For example, $0.5991$ vs. $0.5992$ MSE score on FMNIST for ViLT. Overall, we see that KD induces \textbf{\textit{performance gains across a vast majority of evaluated metrics ($80\%$ for FLAVA and $75\%$ for ViLT)}.} These results indicate that KD in the post-FL phase induces performance improvement across a majority of tasks, retaining high levels of personalization. 

\section{Conclusion and Future Work}
\label{sec:conclusion}
We introduced TAP, a novel two-step adaptive PFL method that enables personalization of heterogeneous multi-modal and multi-task foundation models. TAP is capable of leveraging beneficial knowledge from the server model while maintaining high levels of personalization across clients. We provided convergence analysis to motivate the insufficiency of the server model to cater to the needs of all clients, creating the need for PFL methods such as TAP. Through margin hyperparameters in the FL training period and a KD-based post-FL training period, we demonstrated that TAP possesses superior personalization capabilities across a multitude of datasets, tasks, and model architectures. Avenues of future work could be to explore non-loss based methods of determining replacement (e.g., gradient directions). Moreover, how to set multiple margin values when the number of tasks per client is large remains open for exploration.




\bibliographystyle{plain}
\bibliography{references}
\startcontents[sections]


\newpage
\appendix
\begin{center}
    {\bf\Large Appendix}
\end{center}

\startcontents[sections]
\printcontents[sections]{l}{1}{\setcounter{tocdepth}{3}}

\newpage

\section{LLM Usage}\label{end:llm-usage}
We employed the GPT-5 version of ChatGPT to improve the wording, rephrasing, and overall readability of the manuscript. In addition, it was used as an aid in the creation of tables throughout the manuscript and code implementation. All scientific ideation was produced entirely by the authors.

\section{Technical Details}

\subsection{Pseudocode of TAP}
\label{appendix:pseudocode}
Here, we give a detailed step-by-step pseudocode for the TAP algorithm. The for-loop encompassing global rounds $0$ to $T-1$ entails the FL training protocol and the replacement stage of the algorithm (Sec.~\ref{sec:tap-method}). The last three lines of the algorithm delineate the knowledge-distillation (KD) process after FL communication ends (Sec.~\ref{sec:tap-KD-step}).

\begin{algorithm}[H]
\caption{TAP Training Algorithm}
\label{algo:todo}
\begin{algorithmic}[1]
\Require Clients $c_i \in \mathcal{C}$; Server $S$; datasets $\mathcal{D}_i \subseteq \mathcal{D}$; trainable parameters of local models $\widetilde{\mathbf{W}}_{[i]}$; personalized parameters $\mathbf{X}_{[i]}$; history values $h_{i,o}^{(l)}$ and $h_{i,o}^{(p)}$
\For{$t=0$ \textbf{to} $T-1$}
    \State Sample subset of clients $c_i \in \mathcal{C}_{t}$
    \ParFor{client $c_i \in \mathcal{C}_{t}$}
        \State Train $\widetilde{\mathbf{W}}_{[i]}$ on $\mathcal{D}_i$ with learning rate $\eta_{t}$
        \For{\text{seen task} $o \in \mathcal{O}_{t,i}^{\text{seen}}$}
            \State Update $\widetilde{\mathbf{W}}_{[i]}$'s history value: $h_{i,o}^{(l)} = \ell_{i,o}^{(l)}$
        \EndFor
        \State Transmit $\widetilde{\mathbf{W}}_{[i]}$ from client $c_i$ to server $S$ for aggregation.
        \For{\text{task} $o \in \mathcal{O}_{t,i}^{\text{seen}}$}
            \State Set $\mathbf{R}_i [o]$ to $0$ or $1$ based on \eqref{eq:replacement}
            \State Replacement on $\mathbf{X}_{[i,o]} \subseteq \mathbf{X}_{[i]}$ for task $o$ via \eqref{eq:replacement-final}.
        \EndFor
        \State Reset all entries of $\mathbf{R}_i$ to $0$.
        \State Train $\mathbf{X}_{[i]}$ on $\mathcal{D}_i$ with learning rate $\eta_{t}$
        \For{\text{seen task} $o \in \mathcal{O}_{t,i}^{\text{seen}}$}
            \State Update $\mathbf{X}_{[i]}$'s history value: $h_{i,o}^{(p)} = \ell_{i,o}^{(p)}$
        \EndFor
    \EndParFor
    \State Server $S$ broadcasts $\widetilde{\mathbf{W}}_{[i]}$ to all clients $c_i \in \mathcal{C}$
\EndFor
\For{client $c_i \in \mathcal{C}$}
    \State Train $\widetilde{\mathbf{W}}_{[i]}$ for $P$ minibatch iterations on $\mathcal{D}_i$
    \State Train $\mathbf{X}_{[i]}$ for $P$ minibatch iterations on $\mathcal{D}_i$ utilizing loss from \eqref{eq:kd-loss}.
\EndFor
\end{algorithmic}
\end{algorithm}

\subsection{Server and Local Model Layout}
\label{appendix:model-specifics}
In this section, we present more detailed specifics on the layout of the model structure at server $S$ as seen in Fig.~\ref{fig:model_architecture}. 

As discussed in Sec.~\ref{sec:setup}, we adopt the architecture from \cite{chen2024disentanglement}, with the server model's parameter vector $\mathbf{W} \in \mathbb{R}^{d \times 1}$ holding all modality encoders $\mathbf{W}^{(E)} \in \mathbb{R}^{d^{(E)} \times 1}$ and task decoders $\mathbf{W}^{(D)} \in \mathbb{R}^{d^{(D)} \times 1}$.
The transformer backbone $\mathbf{W}^{(R)} \in \mathbb{R}^{d^{(R)} \times 1}$ sits between the encoders and decoders and consists of two stacks of layers--the first stack, of which there are $|\mathcal{M}+1|$ of for each modality, is responsible for taking the outputs of a modality encoder and routing it as input to the proper layers responsible for that modality. Then, within those layers, the feed forward networks (FFN) are replaced with an MoE, which will activate the expert specialized for the task that the input data is seeking to solve. Then it is routed to the second stack of layers, of which there are $|\mathcal{O}+1|$ of, where it selects the layers responsible for a certain task, utilizing an expert specialized for a certain modality. Then the output is combined and sent to the appropriate task decoder. There are $|\mathcal{M}+1|$ and $|\mathcal{O}+1|$ stacks because each side of the transformer has what are called ``shared" layers, which are activated at all times, no matter the input. The stacks are called the Mixture of Modality Task Expert (MoTE) and Mixture of Modality Expert (MoME) layers respectively.

The combined local model is defined as $\mathbf{W}_{[i]} = \mathbf{W}_{[i]}^{(E)} \cup \mathbf{W}_{[i]}^{(D)} \cup \mathbf{W}_{[i]}^{(R)} \cup \mathbf{A}_i \cup \mathbf{B}_i$ with a learning rate of $\eta_t$. We define the frozen and trainable components of the model for each client $c_i$ as $\widehat{\mathbf{W}}_{[i]} = \mathbf{W}_{[i]}^{(E)} \cup \mathbf{W}_{[i]}^{(R)}$ and $\widetilde{\mathbf{W}}_{[i]} = \mathbf{A}_i \cup \mathbf{B}_i \cup \mathbf{W}_{[i]}^{(D)}$ respectively. For the global model, it is $\widehat{\mathbf{W}} = \mathbf{W}^{(E)} \cup \mathbf{W}^{(R)}$ and $\widetilde{\mathbf{W}} = \mathbf{A} \cup \mathbf{B} \cup \mathbf{W}^{(D)}$. 

\subsection{Memory Footprint and Post-FL Training Time}

\begin{table}[ht]
\centering
\caption{Memory footprint of utilizing TAP vs. no usage of TAP and the wall clock time in seconds of the KD post-FL phase on average for each client.}
\resizebox{\linewidth}{!}{%
\label{table:memory-footprint}
\begin{tabular}{l l c c c c}
\toprule
\textbf{Model} & \textbf{Setting} & \textbf{Avg. Total Params.} & \textbf{Avg. \% trainable} & \textbf{Avg. Memory} & \textbf{Avg. KD Training Time (sec.)} \\
\midrule

\multirow{1}{*}{\centering\textbf{FLAVA}}
& No TAP & $310,559,996.40$ & $10.3347\%$ & $1.1569$ GB & --- \\
& TAP & $347,422,994.40$ & $18.1086\%$ & $1.2943$ GB & $153.54$ \\
\midrule
\multirow{1}{*}{\centering\textbf{ViLT}}
& No TAP & $186,171,919$ & $18.1841\%$ & $0.6935$ GB & --- \\
& TAP & $224,296,094$ & $29.7824\%$ & $0.8357$ GB & $79.39$ \\
\bottomrule
\end{tabular}%
}
\end{table}

Due to the necessity of TAP requiring a post-FL knowledge distillation (Stage 2 of Fig. \ref{fig:two-stage_method}) phase and additional personalized parameters, we seek to characterize how these aspects affects the memory footprint and total extra training time per-client, which can be found in Table~\ref{table:memory-footprint}.

We firstly note that although TAP adds additional parameters, this expansion remains small relative to the full model size. For example, with FLAVA, the total parameter count increases from 310.6M to 347.4M, with ViLT being 186.17M to 224.3M. For TAP, this accounts for about 18\% and 29\% of total parameters, showing that TAP still remains faithful to the idea that fine-tuning foundation models should only involve a small number of the total parameters. Moreover, the memory overhead introduced by TAP is minimal. FLAVA's memory usage increases from 1.16GB to 1.29GB, and ViLT's from 0.69GB to 0.84GB, amounting to a roughly 10-20\% increase. This indicates that the additional personalized parameters do not substantially affect the memory footprint. In addition, the increase in memory footprint outlined is consistent with existing PFL works \cite{sun2023fedperfix}. FedDAT also requires personalized and regular parameters, similar to TAP, which would lead to a similar memory increase. Therefore, \textit{TAP's memory footprint is not inconsistent to existing methodologies in the literature.}

In terms of average per-client knowledge distillation (KD) time, we note that the post-FL KD step adds only a modest amount of computation. The average per-client KD time is approximately 121 seconds, meaning it does not require each client to train for long durations after FL has completed.

\section{Additional Experiments}
\label{appendix:additional-exp}

In this section, we present some experimental results in addition to those presented in Sec.~\ref{exp:main-results}. Similar to Sec.~\ref{exp:main-results}, results outlined in bold signify the highest performing method, with the second best underlined.

\subsection{Number of Replacements}\label{appendix:num-replacements}

\label{figure:replacement}
\begin{figure*}[htbp]
    \centering
    \includegraphics[width=1.0\linewidth]{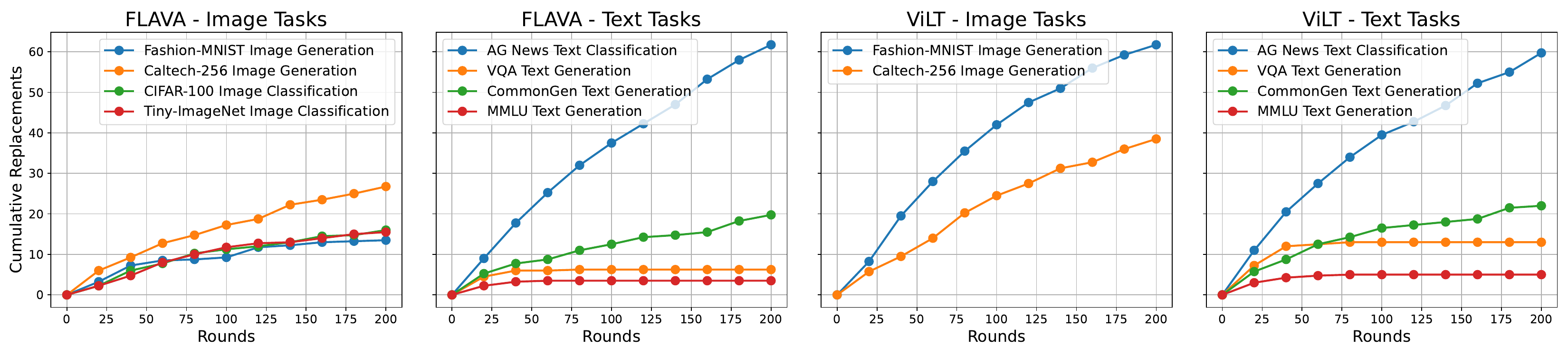}
    \caption{Number of replacements over FL training on FLAVA and ViLT models.}
    \label{fig:replacements}
\end{figure*}

In Fig. \ref{fig:replacements}, we present the number of cumulative replacements taken place over FL training. We note that for a majority of the tasks, most of the replacements take place in the earlier stages of training and then indicate a trend toward leveling off, signifying that reliance on the FL-trained model is most beneficial when the models are still focusing on the basic structure~\citep{frankle2020early}. We note that for certain tasks, such as image generation based tasks on ViLT, as outlined in Table \ref{table:main-img-combined}, slightly under performs compared to some baselines, which means more replacements take place in comparison with other tasks. This indicates that TAP's margin-based method of replacement is an effective means of identifying when replacement is useful, as the $\widetilde{\mathbf{W}}_{[i]}$ model for certain tasks regularly exhibits beneficial information for $\mathbf{X}_{[i]}$.

\subsection{Ablation on Margin Hyperparameters}
\label{appendix:margin-exp}

\begin{table}[ht]
\centering
\caption{Ablation study on margin hyperparameters $m_{i,o}$ for image datasets on ViLT.}
\label{table:margin-vilt-image}
\begin{tabular}{llccc}
\toprule
\textbf{Model} & \textbf{Margin} & \textbf{FMNIST} & \textbf{Caltech-256} & \textbf{Avg. Gen.} \\
\cmidrule(lr){3-5}
 & & \textbf{MSE} $(\downarrow)$ & \textbf{MSE} $(\downarrow)$ & \textbf{MSE} $(\downarrow)$ \\
\midrule

\multirow{3}{*}{\textbf{ViLT}}
& $0.01$ / $0.04$ & $0.6016 \scriptstyle\pm 0.0122$ & $0.5360 \scriptstyle\pm 0.0283$ & $0.5688$ \\
& $0.02$ / $0.05$ & $0.6071 \scriptstyle\pm 0.0132$ & $0.5436 \scriptstyle\pm 0.0334$ & $0.5754$ \\
& $0.05$ / $0.1$ & $0.6384 \scriptstyle\pm 0.0311$ & $0.5783 \scriptstyle\pm 0.0481$ & $0.6084$ \\
\bottomrule
\end{tabular}
\end{table}

\begin{table}[ht]
\centering
\caption{Ablation study on margin hyperparameters $m_{i,o}$ for image datasets on FLAVA.}
\resizebox{\linewidth}{!}{%
\label{table:margin-flava-image}
\begin{tabular}{l l c c c c c c}
\toprule
\textbf{Model} & \textbf{Margin} & \textbf{Tiny-ImageNet} & \textbf{CIFAR-100} & \textbf{FMNIST} & \textbf{Caltech-256} & \textbf{Avg. Class.} & \textbf{Avg. Gen.} \\
\cmidrule(lr){3-8}
 & & \textbf{Acc} $(\uparrow)$ & \textbf{Acc} $(\uparrow)$ & \textbf{MSE} $(\downarrow)$ & \textbf{MSE} $(\downarrow)$ & \textbf{Acc} $(\uparrow)$ & \textbf{MSE} $(\downarrow)$ \\
\midrule

\multirow{3}{*}{\textbf{FLAVA}}
& $0.01$ / $0.04$ & $47.05 \scriptstyle\pm 6.81$ & $56.66 \scriptstyle\pm 4.50$ & $0.5495 \scriptstyle\pm 0.0080$ & $0.3970 \scriptstyle\pm 0.0331$ & $51.85$ & $0.4733$ \\
& $0.02$ / $0.05$ & $47.05 \scriptstyle\pm 6.81$ & $56.24 \scriptstyle\pm 5.49$ & $0.5560 \scriptstyle\pm 0.0110$ & $0.3968 \scriptstyle\pm 0.0313$ & $51.65$ & $0.4764$ \\
& $0.05$ / $0.1$ & $46.78 \scriptstyle\pm 6.54$ & $55.05 \scriptstyle\pm 5.11$ & $0.5674 \scriptstyle\pm 0.0095$ & $0.4354 \scriptstyle\pm 0.0666$ & $50.91$ & $0.5014$ \\
\bottomrule
\end{tabular}%
}
\end{table}

\begin{table}[ht]
\centering
\caption{Ablation study on margin hyperparameters $m_{i,o}$ for text datasets.}
\resizebox{\linewidth}{!}{%
\label{table:margin-text}
\begin{tabular}{l l c cc cc cc cc}
\toprule
\textbf{Model} & \textbf{Margin} & \textbf{AG News} & \multicolumn{2}{c}{\textbf{MMLU}} & \multicolumn{2}{c}{\textbf{VQA}} & \multicolumn{2}{c}{\textbf{CommonGen}} & \multicolumn{2}{c}{\textbf{Avg. Gen.}} \\
\cmidrule(lr){3-11}
& & \textbf{Acc} $(\uparrow)$ & \textbf{BS} $(\uparrow)$ & \textbf{METEOR} $(\uparrow)$ & \textbf{BS} $(\uparrow)$ & \textbf{METEOR} $(\uparrow)$ & \textbf{BS} $(\uparrow)$ & \textbf{METEOR} $(\uparrow)$ & \textbf{BS} $(\uparrow)$ & \textbf{METEOR} $(\uparrow)$ \\
\midrule

\multirow{3}{*}{\centering\textbf{FLAVA}}
& $0.01$ / $0.04$ & $92.73 \scriptstyle\pm 0.31$ & $46.75 \scriptstyle\pm 5.58$ & $21.93 \scriptstyle\pm 4.77$ & $69.62 \scriptstyle\pm 6.83$ & $48.74 \scriptstyle\pm 14.46$ & $35.81 \scriptstyle\pm 3.03$ & $15.32 \scriptstyle\pm 2.60$ & $51.52$ & $30.01$ \\
& $0.02$ / $0.05$ & $92.56 \scriptstyle\pm 0.40$ & $46.75 \scriptstyle\pm 5.58$ & $21.93 \scriptstyle\pm 4.77$ & $69.62 \scriptstyle\pm 6.83$ & $48.74 \scriptstyle\pm 14.46$ & $35.93 \scriptstyle\pm 3.62$ & $15.47 \scriptstyle\pm 2.50$ & $51.57$ & $30.07$ \\
& $0.05$ / $0.1$ & $92.12 \scriptstyle\pm 0.40$ & $46.75 \scriptstyle\pm 5.58$ & $21.93 \scriptstyle\pm 4.77$ & $69.62 \scriptstyle\pm 6.83$ & $48.74 \scriptstyle\pm 14.46$ & $36.23 \scriptstyle\pm 3.32$ & $16.01 \scriptstyle\pm 3.14$ & $51.69$ & $30.29$ \\
\midrule

\multirow{3}{*}{\centering\textbf{ViLT}}
& $0.01$ / $0.04$ & $58.29 \scriptstyle\pm 3.60$ & $57.83 \scriptstyle\pm 3.87$ & $40.74 \scriptstyle\pm 1.04$ & $77.80 \scriptstyle\pm 3.95$ & $60.53 \scriptstyle\pm 12.60$ & $41.79 \scriptstyle\pm 1.16$ & $24.73 \scriptstyle\pm 2.69$ & $59.40$ & $42.25$ \\
& $0.02$ / $0.05$ & $58.22 \scriptstyle\pm 3.39$ & $57.83 \scriptstyle\pm 3.87$ & $40.74 \scriptstyle\pm 1.04$ & $77.80 \scriptstyle\pm 3.95$ & $60.53 \scriptstyle\pm 12.60$ & $41.83 \scriptstyle\pm 1.13$ & $24.92 \scriptstyle\pm 2.66$ & $59.42$ & $42.33$ \\
& $0.05$ / $0.1$ & $57.56 \scriptstyle\pm 4.22$ & $57.83 \scriptstyle\pm 3.87$ & $40.74 \scriptstyle\pm 1.04$ & $77.17 \scriptstyle\pm 4.30$ & $58.99 \scriptstyle\pm 13.40$ & $41.63 \scriptstyle\pm 0.79$ & $25.49 \scriptstyle\pm 2.81$ & $59.09$ & $41.94$ \\
\bottomrule
\end{tabular}%
}
\end{table}

\begin{figure*}[!ht] 
    \centering
    \begin{subfigure}{0.85\linewidth}
        \centering
        \includegraphics[width=\linewidth]{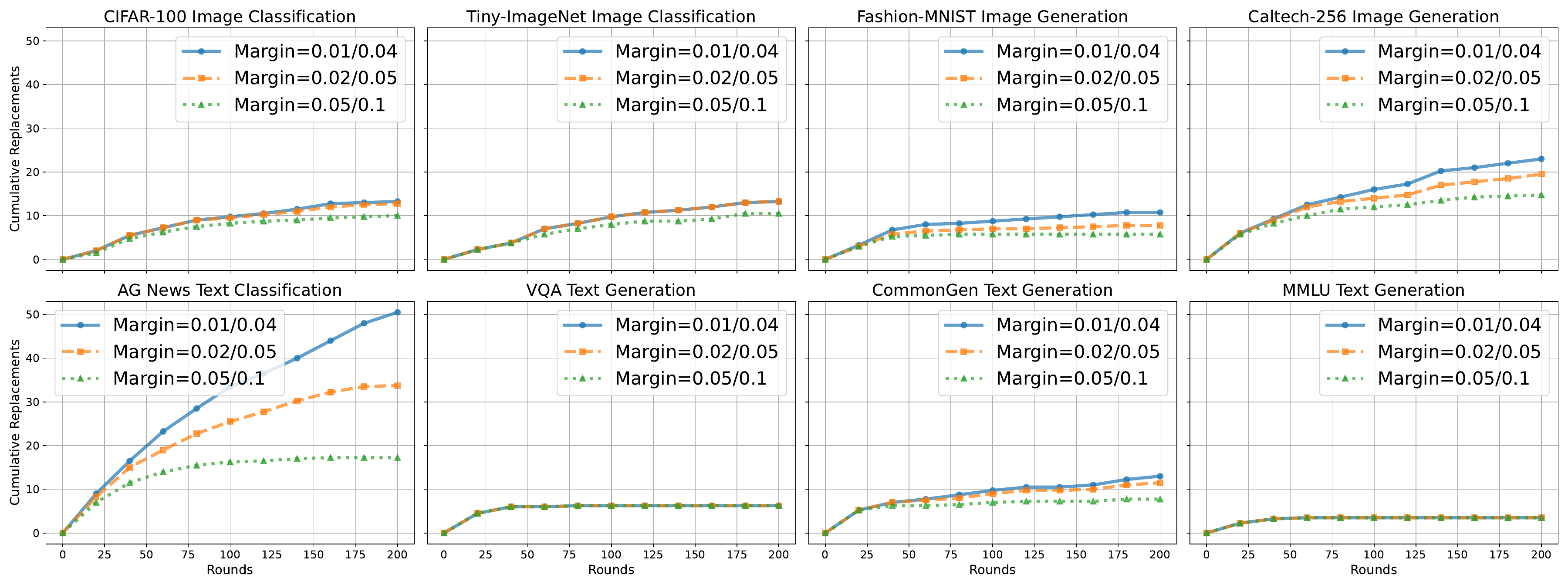}
        \caption{Number of replacements for differing margin settings on FLAVA.}
        \label{fig:flava-margin-replacements}
    \end{subfigure}
    \begin{subfigure}{0.85\linewidth}
        \centering
        \includegraphics[width=\linewidth]{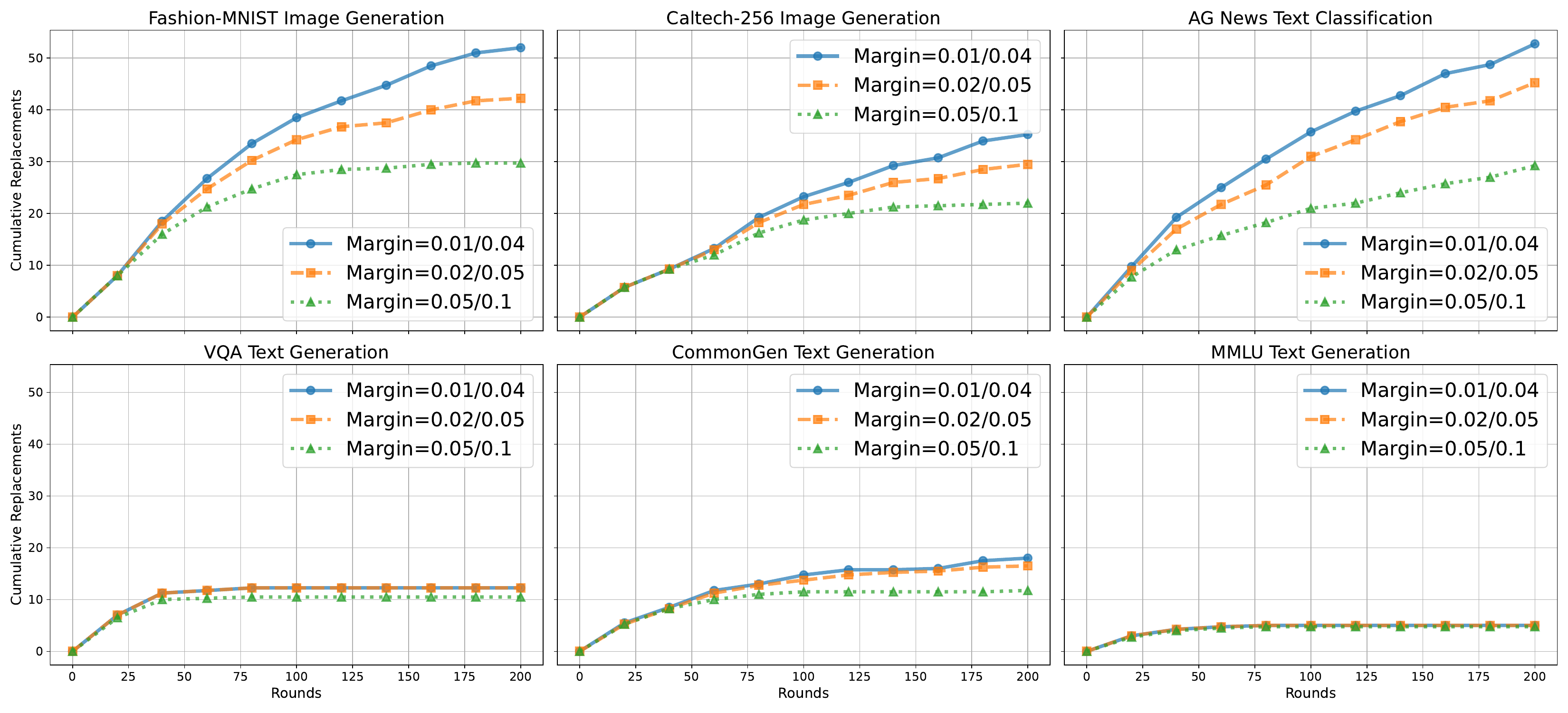}
        \caption{Number of replacements for differing margin settings on ViLT.}
        \label{fig:vilt-margin-replacements}
    \end{subfigure}
    \caption{Replacement comparisons for differing margins on each task across FLAVA and ViLT.}
    \label{fig:margin-replacements}
\end{figure*}

Next, we consider an ablation study on how the margin hyperparameters $m_{i,o}$ influences the performance of TAP. Similar to the style outlined in Table \ref{table:hyperparams-setup-margin}, we consider two differing margin values for each setting depending on the task. For each column, the lower margin value (e.g., 0.01 in the 0.01 / 0.04 column) apply to tasks pertaining to image generation and text classification. The higher margin value is used for all other tasks. Results are given in Tables \ref{table:margin-vilt-image}, \ref{table:margin-flava-image}, and \ref{table:margin-text}.

Based off the results outlined in Tables \ref{table:margin-vilt-image} and \ref{table:margin-flava-image}, we see that image-aligned tasks benefit from having lower $m_{i,o}$ values. For example, the average image classification and generation performance for $m_{i,o}= 0.01$ or $0.04$ is $51.85$ and $0.4733$ while it is $50.91$ and $0.5014$ when $m_{i,o} = 0.05$ or $0.1$ on FLAVA. This means for image-aligned tasks, being more restrictive and setting higher margin values could inhibit $\mathbf{X}_{[i]}$ from regularly taking advantage of the parameters of $\widetilde{\mathbf{W}}_{[i]}$. However, we note margins are not overly sensitive with image classification (e.g., 46.78 vs. 47.05 on Tiny-ImageNet). In Table \ref{table:margin-text}, which deals with text-aligned tasks, we note that VQA and MMLU almost always maintains identical performance across all margins. In terms of other tasks, we note minimal changes in performance between differing explored margin values. Overall, this indicates that for text-aligned tasks (similarly to image classification) the TAP algorithm is not sensitive to the values set for $m_{i,o}$.

Moreover, we also consider how replacements are affected by the differing margin values explored. Based on the results from Fig. \ref{fig:flava-margin-replacements} (FLAVA) and Fig. \ref{fig:vilt-margin-replacements} (ViLT), we note that when we make the threshold for replacement higher, i.e., larger $m_{i,o}$ values, the number of cumulative replacements generally decreases and levels off quicker in comparison to smaller $m_{i,o}$ margins. For example, with AG News, we see for margin values of either $0.01$ or $0.04$, a linear-like increase of cumulative replacements takes place; with $m_{i,o}$ as either $0.05$ or $0.1$, we observe lower total replacements ($\sim 50$ vs. $\sim 18$ replacements on FLAVA and $\sim 50$ vs. $\sim 30$ replacements on ViLT) and a noticeable leveling off around communication round 100. With VQA, CommonGen, and MMLU, we note that the number of replacements remains almost or completely unchanged across all settings, which corroborates with the results from Table \ref{table:margin-text}, where the BS and METEOR scores remain mostly unchanged across differing margins. Overall, these outcomes indicate that in general, the margin hyperparameters play an important role in shaping $\mathbf{X}_{[i]}$'s interaction with FL-engaged $\widetilde{\mathbf{W}}_{[i]}$.

\subsection{Post-FL training for $2 \cdot P$ iterations vs. TAP}
\label{appendix:2p-exp}
Since the proposed TAP algorithm relies on $\widetilde{\mathbf{W}}_{T, [i]}$ training for $P$ mini-batch iterations after FL before also training $\mathbf{X}_{T, [i]}$ for $P$ iterations with $\widetilde{\mathbf{W}}_{T, [i]}$ as the teacher model, we seek to see if $\mathbf{X}_{T, [i]}$ trained via the KD-based post-FL process still achieves better performance than $\widetilde{\mathbf{W}}_{T, [i]}$ being trained for $2 \cdot P$ iterations. We do not include FedDAT~\citep{chen2024feddat} in this experiment, as its optimization procedure involves sequentially updating the aggregated and personal adapters, resulting in the same number of forward passes as required by TAP during post-FL training. For similar reasons, we also do not include Per-FedAvg~\citep{fallah2020personalized}, as it must conduct two passes and obtain two differing gradients for an inner and outer update step.
\begin{table}[H] 
\centering
\caption{TAP vs. DisentAFL and FedAvg trained for $2P$ iterations on image tasks for ViLT.}
\label{table:2p-vilt-image}
\begin{tabular}{lccc}
\toprule
\textbf{Method (ViLT)} & \textbf{FMNIST} & \textbf{Caltech-256} & \textbf{Avg. Gen.} \\
\cmidrule(lr){2-4}
 & \textbf{MSE} $(\downarrow)$ & \textbf{MSE} $(\downarrow)$ & \textbf{MSE} $(\downarrow)$ \\
\midrule

FedAvg + $2 \cdot P$ & $\underline{0.6019} \scriptstyle\pm 0.0146$ & $\textbf{0.5294} \scriptstyle\pm 0.0291$ & $\underline{0.5657}$ \\
DisentAFL + $2 \cdot P$ & $2.6408 \scriptstyle\pm 1.4097$ & $2.5918 \scriptstyle\pm 1.9660$ & $2.6163$ \\
\rowcolor{blue!10}
\textbf{TAP (Ours)} & $\textbf{0.5992} \scriptstyle\pm 0.0117$ & $\underline{0.5300} \scriptstyle\pm 0.0247$ & $\textbf{0.5646}$ \\
\bottomrule
\end{tabular}
\end{table}

\begin{table}[H]
\centering
\caption{TAP vs. DisentAFL and FedAvg trained for $2P$ iterations on image tasks for FLAVA.}
\resizebox{\linewidth}{!}{%
\label{table:2p-flava-image}
\begin{tabular}{l c c c c c c}
\toprule
\textbf{Method (FLAVA)} & \textbf{Tiny-ImageNet} & \textbf{CIFAR-100} & \textbf{FMNIST} & \textbf{Caltech-256} & \textbf{Avg. Class.} & \textbf{Avg. Gen.} \\
\cmidrule(lr){2-7}
 & \textbf{Acc} $(\uparrow)$ & \textbf{Acc} $(\uparrow)$ & \textbf{MSE} $(\downarrow)$ & \textbf{MSE} $(\downarrow)$ & \textbf{Acc} $(\uparrow)$ & \textbf{MSE} $(\downarrow)$ \\
\midrule

FedAvg + $2 \cdot P$ & $44.88 \scriptstyle\pm 7.37$ & $\underline{56.14} \scriptstyle\pm 4.97$ & $\underline{0.5498} \scriptstyle\pm 0.0065$ & $\textbf{0.3896} \scriptstyle\pm 0.0287$ & $\underline{50.51}$ & $\underline{0.4697}$ \\
DisentAFL + $2 \cdot P$ & $\textbf{48.84} \scriptstyle\pm 7.47$ & $7.26 \scriptstyle\pm 1.00$ & $0.6904 \scriptstyle\pm 0.0302$ & $0.5192 \scriptstyle\pm 0.0382$ & $28.05$ & $0.6048$ \\
\rowcolor{blue!10}
\textbf{TAP (Ours)} & $\underline{47.30} \scriptstyle\pm 6.57$ & $\textbf{56.84} \scriptstyle\pm 4.70$ & $\textbf{0.5478} \scriptstyle\pm 0.0074$ & $\underline{0.3899} \scriptstyle\pm 0.0305$ & $\textbf{52.07}$ & $\textbf{0.4689}$ \\
\bottomrule
\end{tabular}%
}
\end{table}

\begin{table}[H]
\centering
\caption{TAP vs. DisentAFL and FedAvg trained for $2P$ iterations on text tasks.}
\resizebox{\linewidth}{!}{%
\label{table:2p-text}
\begin{tabular}{l l c cc cc cc cc}
\toprule
\textbf{Model} & \textbf{Method} & \textbf{AG News} & \multicolumn{2}{c}{\textbf{MMLU}} & \multicolumn{2}{c}{\textbf{VQA}} & \multicolumn{2}{c}{\textbf{CommonGen}} & \multicolumn{2}{c}{\textbf{Avg. Gen.}} \\
\cmidrule(lr){3-11}
& & \textbf{Acc} $(\uparrow)$ & \textbf{BS} $(\uparrow)$ & \textbf{METEOR} $(\uparrow)$ & \textbf{BS} $(\uparrow)$ & \textbf{METEOR} $(\uparrow)$ & \textbf{BS} $(\uparrow)$ & \textbf{METEOR} $(\uparrow)$ & \textbf{BS} $(\uparrow)$ & \textbf{METEOR} $(\uparrow)$ \\
\midrule

\multirow{3}{*}{\centering\textbf{FLAVA}}
& FedAvg + $2 \cdot P$ & $\textbf{93.07} \scriptstyle\pm 0.16$ & $39.42 \scriptstyle\pm 2.21$ & $13.02 \scriptstyle\pm 4.98$ & $\underline{53.36} \scriptstyle\pm 5.24$ & $\underline{30.37} \scriptstyle\pm 8.29$ & $\underline{31.73} \scriptstyle\pm 1.63$ & $11.86 \scriptstyle\pm 1.61$ & $\underline{41.92}$ & $\underline{19.50}$ \\
& DisentAFL + $2 \cdot P$ & $\underline{92.84} \scriptstyle\pm 0.23$ & $\underline{40.15} \scriptstyle\pm 0.42$ & $\underline{20.12} \scriptstyle\pm 1.76$ & $40.69 \scriptstyle\pm 2.11$ & $23.00 \scriptstyle\pm 3.69$ & $29.32 \scriptstyle\pm 1.19$ & $\underline{13.96} \scriptstyle\pm 1.61$ & $36.03$ & $18.81$ \\
\rowcolor{blue!10}
& \textbf{TAP (Ours)} & $92.83 \scriptstyle\pm 0.26$ & $\textbf{46.75} \scriptstyle\pm 5.58$ & $\textbf{21.93} \scriptstyle\pm 4.77$ & $\textbf{69.62} \scriptstyle\pm 6.83$ & $\textbf{48.74} \scriptstyle\pm 14.46$ & $\textbf{35.61} \scriptstyle\pm 3.34$ & $\textbf{15.12} \scriptstyle\pm 2.60$ & $\textbf{51.44}$ & $\textbf{29.93}$ \\
\midrule

\multirow{3}{*}{\centering\textbf{ViLT}}
& FedAvg + $2 \cdot P$ & $\underline{58.40} \scriptstyle\pm 3.33$ & $46.56 \scriptstyle\pm 4.45$ & $12.43 \scriptstyle\pm 9.21$ & $\underline{58.66} \scriptstyle\pm 12.47$ & $\underline{44.01} \scriptstyle\pm 22.08$ & $\underline{38.36} \scriptstyle\pm 2.43$ & $\underline{13.06} \scriptstyle\pm 7.14$ & $\underline{48.12}$ & $\underline{25.31}$ \\
& DisentAFL + $2 \cdot P$ & $57.15 \scriptstyle\pm 3.46$ & $\underline{54.92} \scriptstyle\pm 0.58$ & $\underline{40.67} \scriptstyle\pm 0.56$ & $46.58 \scriptstyle\pm 1.95$ & $25.25 \scriptstyle\pm 1.73$ & $35.19 \scriptstyle\pm 0.54$ & $9.95 \scriptstyle\pm 0.61$ & $43.69$ & $22.22$ \\
\rowcolor{blue!10}
& \textbf{TAP (Ours)} & $\textbf{58.42} \scriptstyle\pm 3.52$ & $\textbf{57.83} \scriptstyle\pm 3.87$ & $\textbf{40.74} \scriptstyle\pm 1.04$ & $\textbf{77.80} \scriptstyle\pm 3.95$ & $\textbf{60.53} \scriptstyle\pm 12.60$ & $\textbf{41.93} \scriptstyle\pm 1.05$ & $\textbf{25.54} \scriptstyle\pm 1.46$ & $\textbf{59.46}$ & $\textbf{42.58}$ \\
\bottomrule
\end{tabular}%
}
\end{table}

Based off the results of Tables \ref{table:2p-vilt-image}, \ref{table:2p-flava-image}, and \ref{table:2p-text}, we see that across a majority of tasks, TAP still outperforms both FedAvg + $2 \cdot P$ and DisentAFL + $2 \cdot P$. In addition, we see that when either FedAvg + $2 \cdot P$ or DisentAFL + $2 \cdot P$ is better in performance, TAP always remains close to the best performing baseline, as seen with examples of $0.5300$ (TAP) vs. $0.5294$ (FedAvg + $2 \cdot P$) for Caltech-256 on ViLT and $92.83$ (TAP) vs. $93.07$ (FedAvg + $2 \cdot P$) for AG News on FLAVA. This demonstrates that the final personalized model produced by TAP induces greater levels of personalization in comparison to merely devoting more iterations to $\widetilde{\mathbf{W}}_{T, [i]}$.

\subsection{TAP vs. Standard Transformer Architecture}

\begin{table}[ht]
\centering
\caption{Standard transformer backbone with baselines vs. TAP on image datasets for ViLT.}
\scriptsize
\setlength{\tabcolsep}{4pt} 
\renewcommand{\arraystretch}{0.9} 
\label{table:unified-vilt-img}
\begin{tabular}{l c c c}
\toprule
\textbf{Method (ViLT)} & \textbf{FMNIST} & \textbf{Caltech-256} & \textbf{Avg. Gen.} \\
\cmidrule(lr){2-4}
 & \textbf{MSE} $(\downarrow)$ & \textbf{MSE} $(\downarrow)$ & \textbf{MSE} $(\downarrow)$ \\
\midrule
FedAvg & $0.8230 \scriptstyle\pm 0.1037$ & $0.8677 \scriptstyle\pm 0.0659$ & $0.8454$ \\
FedAvg + Post-train & $0.8182 \scriptstyle\pm 0.1025$ & $0.8611 \scriptstyle\pm 0.0657$ & $0.8397$ \\
Per-FedAvg & $0.8381 \scriptstyle\pm 0.0973$ & $0.9156 \scriptstyle\pm 0.0585$ & $0.8769$ \\
Per-FedAvg + Post-train & $\underline{0.8165} \scriptstyle\pm 0.1019$ & $\underline{0.8548} \scriptstyle\pm 0.0659$ & $\underline{0.8357}$ \\
FedDAT & $0.8224 \scriptstyle\pm 0.1024$ & $0.8720 \scriptstyle\pm 0.0682$ & $0.8472$ \\
FedDAT + Post-train & $0.8184 \scriptstyle\pm 0.1011$ & $0.8675 \scriptstyle\pm 0.0679$ & $0.8430$ \\
\rowcolor{blue!10}
\textbf{TAP (Ours)} & $\mathbf{0.5992} \scriptstyle\pm 0.0117$ & $\mathbf{0.5300} \scriptstyle\pm 0.0247$ & $\mathbf{0.5646}$ \\
\bottomrule
\end{tabular}
\end{table}

\begin{table}[ht]
\centering
\caption{Standard transformer backbone with baselines vs. TAP on image datasets for FLAVA.}
\resizebox{\linewidth}{!}{%
\label{table:unified-flava-image}
\begin{tabular}{l c c c c c c}
\toprule
\textbf{Method (FLAVA)} & \textbf{Tiny-ImageNet} & \textbf{CIFAR-100} & \textbf{FMNIST} & \textbf{Caltech-256} & \textbf{Avg. Class.} & \textbf{Avg. Gen.} \\
\cmidrule(lr){2-7}
 & \textbf{Acc} $(\uparrow)$ & \textbf{Acc} $(\uparrow)$ & \textbf{MSE} $(\downarrow)$ & \textbf{MSE} $(\downarrow)$ & \textbf{Acc} $(\uparrow)$ & \textbf{MSE} $(\downarrow)$ \\
\midrule

FedAvg & $27.93 \scriptstyle\pm 12.69$ & $35.25 \scriptstyle\pm 12.58$ & $0.6005 \scriptstyle\pm 0.0169$ & $0.6128 \scriptstyle\pm 0.0563$ & $31.59$ & $0.6067$ \\
FedAvg + Post-train & $\underline{29.64} \scriptstyle\pm 12.56$ & $37.00 \scriptstyle\pm 12.59$ & $0.5953 \scriptstyle\pm 0.0154$ & $0.5960 \scriptstyle\pm 0.0524$ & $\underline{33.32}$ & $0.5957$ \\
Per-FedAvg & $26.93 \scriptstyle\pm 13.76$ & $35.10 \scriptstyle\pm 12.45$ & $0.7177 \scriptstyle\pm 0.0320$ & $0.8592 \scriptstyle\pm 0.1112$ & $31.02$ & $0.7885$ \\
Per-FedAvg + Post-train & $28.83 \scriptstyle\pm 13.88$ & $\underline{37.69} \scriptstyle\pm 12.31$ & $0.6097 \scriptstyle\pm 0.0195$ & $0.6137 \scriptstyle\pm 0.0474$ & $33.26$ & $0.6117$ \\
FedDAT & $17.65 \scriptstyle\pm 16.77$ & $30.45 \scriptstyle\pm 18.06$ & $0.5817 \scriptstyle\pm 0.0128$ & $0.5650 \scriptstyle\pm 0.0543$ & $24.05$ & $0.5734$ \\
FedDAT + Post-train & $19.15 \scriptstyle\pm 17.14$ & $32.59 \scriptstyle\pm 18.85$ & $\underline{0.5782} \scriptstyle\pm 0.0124$ & $\underline{0.5540} \scriptstyle\pm 0.0518$ & $25.87$ & $\underline{0.5661}$ \\
\rowcolor{blue!10}
\textbf{TAP (Ours)} & $\textbf{47.30} \scriptstyle\pm 6.57$ & $\textbf{56.84} \scriptstyle\pm 4.70$ & $\textbf{0.5478} \scriptstyle\pm 0.0074$ & $\textbf{0.3899} \scriptstyle\pm 0.0305$ & $\textbf{52.07}$ & $\textbf{0.4689}$ \\
\bottomrule
\end{tabular}%
}
\end{table}

\begin{table}[!ht] 
\centering
\caption{Standard transformer backbone with baselines vs. TAP on text datasets.}
\resizebox{\linewidth}{!}{%
\label{table:unified-txt}
\begin{tabular}{l l c cc cc cc cc}
\toprule
\textbf{Model} & \textbf{Method} & \textbf{AG News} & \multicolumn{2}{c}{\textbf{MMLU}} & \multicolumn{2}{c}{\textbf{VQA}} & \multicolumn{2}{c}{\textbf{CommonGen}} & \multicolumn{2}{c}{\textbf{Avg. Gen.}} \\
\cmidrule(lr){3-11}
& & \textbf{Acc} $(\uparrow)$ & \textbf{BS} $(\uparrow)$ & \textbf{METEOR} $(\uparrow)$ & \textbf{BS} $(\uparrow)$ & \textbf{METEOR} $(\uparrow)$ & \textbf{BS} $(\uparrow)$ & \textbf{METEOR} $(\uparrow)$ & \textbf{BS} $(\uparrow)$ & \textbf{METEOR} $(\uparrow)$ \\
\midrule

\multirow{1}{*}{\centering\textbf{FLAVA}}
& FedAvg & $90.48 \scriptstyle\pm 0.60$ & $41.67 \scriptstyle\pm 1.24$ & $21.39 \scriptstyle\pm 2.61$ & $48.32 \scriptstyle\pm 1.91$ & $10.88 \scriptstyle\pm 2.52$ & $30.45 \scriptstyle\pm 1.34$ & $13.07 \scriptstyle\pm 2.41$ & $39.84$ & $13.86$ \\
& FedAvg + Post-train & $\underline{90.78} \scriptstyle\pm 0.49$ & $\underline{41.96} \scriptstyle\pm 1.05$ & $\textbf{22.04} \scriptstyle\pm 1.97$ & $48.06 \scriptstyle\pm 2.00$ & $10.46 \scriptstyle\pm 2.19$ & $30.45 \scriptstyle\pm 1.44$ & $13.05 \scriptstyle\pm 2.46$ & $39.80$ & $13.81$ \\
& Per-FedAvg & $89.64 \scriptstyle\pm 1.45$ & $\underline{41.96} \scriptstyle\pm 2.57$ & $19.42 \scriptstyle\pm 4.88$ & $47.08 \scriptstyle\pm 2.53$ & $10.26 \scriptstyle\pm 2.12$ & $30.81 \scriptstyle\pm 1.53$ & $13.33 \scriptstyle\pm 2.75$ & $39.46$ & $13.47$ \\
& Per-FedAvg + Post-train & $90.67 \scriptstyle\pm 0.51$ & $41.83 \scriptstyle\pm 1.02$ & $21.26 \scriptstyle\pm 1.74$ & $47.32 \scriptstyle\pm 2.38$ & $10.20 \scriptstyle\pm 2.01$ & $30.56 \scriptstyle\pm 1.47$ & $13.46 \scriptstyle\pm 2.75$ & $39.51$ & $13.71$ \\
& FedDAT & $90.21 \scriptstyle\pm 0.68$ & $41.63 \scriptstyle\pm 1.05$ & $21.06 \scriptstyle\pm 1.96$ & $\underline{67.65} \scriptstyle\pm 11.40$ & $\textbf{57.74} \scriptstyle\pm 12.10$ & $\underline{40.81} \scriptstyle\pm 2.29$ & $\textbf{25.58} \scriptstyle\pm 1.17$ & $\textbf{51.71}$ & $\textbf{37.54}$ \\
& FedDAT + Post-train & $90.67 \scriptstyle\pm 0.53$ & $41.70 \scriptstyle\pm 0.81$ & $21.87 \scriptstyle\pm 2.40$ & $67.28 \scriptstyle\pm 11.40$ & $\underline{57.38} \scriptstyle\pm 12.74$ & $\textbf{40.82} \scriptstyle\pm 1.95$ & $\underline{24.80} \scriptstyle\pm 2.46$ & $\underline{51.58}$ & $\underline{37.24}$ \\
\rowcolor{blue!10}
& \textbf{TAP (Ours)} & $\textbf{92.83} \scriptstyle\pm 0.26$ & $\textbf{46.75} \scriptstyle\pm 5.58$ & $\underline{21.93} \scriptstyle\pm 4.77$ & $\textbf{69.62} \scriptstyle\pm 6.83$ & $48.74 \scriptstyle\pm 14.46$ & $35.61 \scriptstyle\pm 3.34$ & $15.12 \scriptstyle\pm 2.60$ & $51.44$ & $29.93$ \\
\midrule

\multirow{1}{*}{\centering\textbf{ViLT}}
& FedAvg & $81.32 \scriptstyle\pm 1.82$ & $55.53 \scriptstyle\pm 1.82$ & $\underline{40.57} \scriptstyle\pm 1.91$ & $\underline{58.89} \scriptstyle\pm 1.79$ & $\underline{28.82} \scriptstyle\pm 5.77$ & $41.30 \scriptstyle\pm 1.27$ & $25.18 \scriptstyle\pm 2.54$ & $\underline{51.18}$ & $29.71$ \\
& FedAvg + Post-train & $81.97 \scriptstyle\pm 1.45$ & $54.86 \scriptstyle\pm 0.98$ & $39.78 \scriptstyle\pm 2.03$ & $58.61 \scriptstyle\pm 2.05$ & $28.75 \scriptstyle\pm 5.92$ & $41.36 \scriptstyle\pm 1.37$ & $\underline{25.96} \scriptstyle\pm 1.76$ & $50.96$ & $\underline{29.84}$ \\
& Per-FedAvg & $73.40 \scriptstyle\pm 5.25$ & $\underline{55.74} \scriptstyle\pm 1.89$ & $40.44 \scriptstyle\pm 2.46$ & $58.01 \scriptstyle\pm 1.87$ & $27.50 \scriptstyle\pm 4.89$ & $40.86 \scriptstyle\pm 1.39$ & $24.65 \scriptstyle\pm 2.56$ & $50.70$ & $28.95$ \\
& Per-FedAvg + Post-train & $81.24 \scriptstyle\pm 1.73$ & $55.26 \scriptstyle\pm 1.04$ & $40.17 \scriptstyle\pm 1.84$ & $58.19 \scriptstyle\pm 1.67$ & $27.83 \scriptstyle\pm 4.45$ & $40.91 \scriptstyle\pm 1.41$ & $25.29 \scriptstyle\pm 1.92$ & $50.69$ & $29.29$ \\
& FedDAT & $\underline{82.58} \scriptstyle\pm 1.59$ & $55.73 \scriptstyle\pm 1.94$ & $40.21 \scriptstyle\pm 2.32$ & $57.93 \scriptstyle\pm 2.52$ & $26.83 \scriptstyle\pm 7.31$ & $41.36 \scriptstyle\pm 1.48$ & $25.44 \scriptstyle\pm 1.96$ & $50.86$ & $28.95$ \\
& FedDAT + Post-train & $\textbf{83.36} \scriptstyle\pm 1.40$ & $55.07 \scriptstyle\pm 1.01$ & $40.06 \scriptstyle\pm 2.05$ & $57.82 \scriptstyle\pm 2.60$ & $26.68 \scriptstyle\pm 6.70$ & $\underline{41.38} \scriptstyle\pm 1.49$ & $\textbf{26.01} \scriptstyle\pm 1.90$ & $50.69$ & $29.09$ \\
\rowcolor{blue!10}
& \textbf{TAP (Ours)} & $58.42 \scriptstyle\pm 3.52$ & $\textbf{57.83} \scriptstyle\pm 3.87$ & $\textbf{40.74} \scriptstyle\pm 1.04$ & $\textbf{77.80} \scriptstyle\pm 3.95$ & $\textbf{60.53} \scriptstyle\pm 12.60$ & $\textbf{41.93} \scriptstyle\pm 1.05$ & $25.54 \scriptstyle\pm 1.46$ & $\textbf{59.46}$ & $\textbf{42.58}$ \\
\bottomrule
\end{tabular}%
}
\end{table}

Here, we compare the usage of TAP with the model architecture adopted from \cite{chen2024disentanglement} against a more standard architecture with a non modality-task pair-aware server for the baselines. In this scenario, each client employs a standard transformer layer structure, with no MoE routing taking place (i.e., each client shares the same backbone), which is sent to the server for vanilla FedAvg (not component-wise) aggregation~\citep{mcmahan2017communication}. 

From Tables \ref{table:unified-vilt-img}, \ref{table:unified-flava-image}, and \ref{table:unified-txt}, we note that TAP outperforms the baselines on a majority of the tasks evaluated. Even when TAP lags behind one of the baselines, such as with TAP vs. FedDAT on CommonGen, it is the most consistent method across all tasks. For example, TAP retains an average BS score and image accuracy of 51.44 and 52.07 respectively on FLAVA. By contrast, FedDAT + Post-train obtains 51.58 and 25.87, indicating that while FedDAT + Post-train achieves marginally better average BS scores, it significantly underperforms TAP on image classification. Overall, this demonstrates the merit of adopting a model architecture that is modality-task pair aware, to better prevent conflicting tasks from competing with one another and ensuring better consistency in performance across all tasks.

\subsection{Statistical Significance of TAP's Improvement vs. Baselines}
\label{appendix:welch-t-test}

\begin{table}[ht]
\centering
\caption{Percentage of metrics for which TAP exhibits superior performance and is statistically significant with a significance threshold of $0.05$ in comparison to each baseline.}
\label{table:welch-t-test}
\begin{tabular}{llc}
\toprule
\textbf{Model} & \textbf{Baseline} & \textbf{\% Metrics TAP Significantly Better} \\
\midrule
\multirow{9}{*}{\textbf{FLAVA}}
& Local & 90.9\% \\
& FedAvg & 90.9\% \\
& FedAvg + Post-train & 63.6\% \\
& Per-FedAvg & 100.0\% \\
& Per-FedAvg + Post-train & 63.6\% \\
& DisentAFL & 81.8\% \\
& DisentAFL + Post-train & 72.7\% \\
& FedDAT & 72.7\% \\
& FedDAT + Post-train & 45.5\% \\
\midrule
\multirow{8}{*}{\textbf{ViLT}}
& Local & 55.6\% \\
& FedAvg & 88.9\% \\
& FedAvg + Post-train & 77.8\% \\
& Per-FedAvg & 100.0\% \\
& Per-FedAvg + Post-train & 88.9\% \\
& DisentAFL & 88.9\% \\
& DisentAFL + Post-train & 77.8\% \\
& FedDAT & 88.9\% \\
& FedDAT + Post-train & 66.7\% \\
\bottomrule
\end{tabular}
\end{table}

In this section, we seek to see if the improvements exhibited from TAP in Tables \ref{table:main-img-combined} and \ref{table:main-txt} from Sec. \ref{exp:main-results} is statistically significant. In Table~\ref{table:welch-t-test}, we utilize the commonly used Welch $t$-test~\cite{west2021best} with a standard significance threshold of $0.05$ to see the percentage of metrics for which TAP is both (1) the higher performing method and (2) statistically significant.

From Table~\ref{table:welch-t-test}, we note that across nearly all baselines and settings, TAP achieves statistically significant improvements on the majority of metrics. Even in cases where significance is comparatively lower (e.g., FedDAT + Post-train), TAP remains consistently stronger across architectures. These findings give credence that TAP's gains are not only numerically stronger but statistically reliable.

\subsection{Deeper Backbone Impact}

\begin{table*}[htbp] 
\centering
\caption{Performance comparison for deeper FLAVA (\textasciitilde800M more parameters) on image datasets.}
\resizebox{\linewidth}{!}{%
\label{table:appx-flava-img-deep}
\begin{tabular}{l c c c c c c}
\toprule
\textbf{Method (FLAVA)} & \textbf{Tiny-ImageNet} & \textbf{CIFAR-100} & \textbf{FMNIST} & \textbf{Caltech-256} & \textbf{Avg. Class.} & \textbf{Avg. Gen.} \\
\cmidrule(lr){2-7}
 & \textbf{Acc} $(\uparrow)$ & \textbf{Acc} $(\uparrow)$ & \textbf{MSE} $(\downarrow)$ & \textbf{MSE} $(\downarrow)$ & \textbf{Acc} $(\uparrow)$ & \textbf{MSE} $(\downarrow)$ \\
\midrule

FedAvg & $\underline{14.50}$ & $23.12$ & $0.7619$ & $0.7058$ & $18.81$ & $0.7339$ \\
FedAvg + Post-train & $14.01$ & $\underline{24.73}$ & $0.6871$ & $0.6420$ & $\underline{19.37}$ & $0.6646$ \\
FedDAT & $5.80$ & $17.37$ & $0.7923$ & $0.7317$ & $11.59$ & $0.7620$ \\
FedDAT + Post-train & $6.23$ & $17.77$ & $\underline{0.6779}$ & $\underline{0.6137}$ & $12.00$ & $\underline{0.6458}$ \\
\rowcolor{blue!10}
\textbf{TAP (Ours)} & $\textbf{23.06}$ & $\textbf{34.79}$ & $\textbf{0.6111}$ & $\textbf{0.5846}$ & $\textbf{28.93}$ & $\textbf{0.5979}$ \\
\bottomrule
\end{tabular}%
}
\end{table*}

\begin{table*}[htbp] 
\centering
\caption{Performance comparison for deeper FLAVA (\textasciitilde800M more parameters) on text datasets.}
\resizebox{\linewidth}{!}{%
\label{table:appx-txt-deep}
\begin{tabular}{l c cc cc cc cc}
\toprule
\textbf{Method} & \textbf{AG News} & \multicolumn{2}{c}{\textbf{MMLU}} & \multicolumn{2}{c}{\textbf{VQA}} & \multicolumn{2}{c}{\textbf{CommonGen}} & \multicolumn{2}{c}{\textbf{Avg. Gen.}} \\
\cmidrule(lr){2-10}
& \textbf{Acc} $(\uparrow)$ & \textbf{BS} $(\uparrow)$ & \textbf{METEOR} $(\uparrow)$ & \textbf{BS} $(\uparrow)$ & \textbf{METEOR} $(\uparrow)$ & \textbf{BS} $(\uparrow)$ & \textbf{METEOR} $(\uparrow)$ & \textbf{BS} $(\uparrow)$ & \textbf{METEOR} $(\uparrow)$ \\
\midrule

FedAvg & $92.24$ & $41.21$ & $12.17$ & $32.65$ & $5.97$ & $26.06$ & $7.33$ & $31.73$ & $7.75$ \\
FedAvg + Post-train & $91.19$ & $\textbf{41.31}$ & $12.36$ & $\underline{33.14}$ & $6.59$ & $27.08$ & $7.51$ & $32.35$ & $8.11$ \\
FedDAT & $\textbf{92.33}$ & $39.61$ & $\underline{24.51}$ & $30.04$ & $7.92$ & $26.24$ & $6.05$ & $30.43$ & $10.49$ \\
FedDAT + Post-train & $\underline{92.28}$ & $40.30$ & $\textbf{26.39}$ & $33.02$ & $\underline{9.35}$ & $\textbf{30.07}$ & $\underline{8.17}$ & $\underline{33.29}$ & $\underline{12.29}$ \\
\rowcolor{blue!10}
\textbf{TAP (Ours)} & $92.13$ & $\underline{41.24}$ & $19.21$ & $\textbf{53.80}$ & $\textbf{23.43}$ & $\underline{27.97}$ & $\textbf{10.64}$ & $\textbf{40.95}$ & $\textbf{17.47}$ \\
\bottomrule
\end{tabular}%
}
\vspace{-3mm}
\end{table*}

Next, we consider if TAP still maintains superior performance to the baselines if more layers in the backbone (compared to the main experiments) are considered. Here, we consider a setup where the transformer backbone of FLAVA is deeper, with approximately $800$M more parameters. 

From the presented metrics in Tables \ref{table:appx-flava-img-deep} and \ref{table:appx-txt-deep}, we can note the superiority of TAP across nearly all metrics in comparison to the baselines. Similar to Sec. \ref{exp:main-results}, we note that even when TAP is not the best performing method, it is almost always the next best option and more consistent across tasks and modalities (e.g., $23.06$ vs $6.23$ on Tiny-ImageNet and $92.13$ vs $92.28$ on AG News for TAP and FedDAT + Post-train). This indicates that TAP still maintains higher performance than state-of-the-art baselines on differing backbone sizes.

\subsection{Large Number of Clients}
\begin{table*}[htbp] 
\centering
\caption{Performance comparison for FLAVA on image datasets with 100 clients and 5 clients aggregated each round.}
\resizebox{\linewidth}{!}{%
\label{table:appendix-flava-img-100clients}
\begin{tabular}{l c c c c c c}
\toprule
\textbf{Method (FLAVA)} & \textbf{ImageNet} & \textbf{CIFAR-100} & \textbf{FMNIST} & \textbf{Caltech-256} & \textbf{Avg. Class.} & \textbf{Avg. Gen.} \\
\cmidrule(lr){2-7}
 & \textbf{Acc} $(\uparrow)$ & \textbf{Acc} $(\uparrow)$ & \textbf{MSE} $(\downarrow)$ & \textbf{MSE} $(\downarrow)$ & \textbf{Acc} $(\uparrow)$ & \textbf{MSE} $(\downarrow)$ \\
\midrule

FedAvg & $37.32$ & $47.23$ & $0.6612$ & $0.5446$ & $42.28$ & $0.6029$ \\
FedAvg + Post-train & $43.70$ & $55.15$ & $0.5860$ & $0.5301$ & $\underline{49.43}$ & $\textbf{0.5581}$ \\
\rowcolor{blue!10}
\rowcolor{blue!10}
\textbf{TAP (Ours)} & $45.16$ & $54.73$ & $0.5961$ & $0.5529$ & $\textbf{49.95}$ & $\underline{0.5745}$ \\
\bottomrule
\end{tabular}%
}
\end{table*}

\begin{table*}[htbp] 
\centering
\caption{Performance comparison for FLAVA on text datasets with 100 clients and 5 clients aggregated each round.}
\resizebox{\linewidth}{!}{%
\label{table:appendix-txt-100clients}
\begin{tabular}{l c cc cc cc cc}
\toprule
\textbf{Method} & \textbf{AG News} & \multicolumn{2}{c}{\textbf{MMLU}} & \multicolumn{2}{c}{\textbf{VQA}} & \multicolumn{2}{c}{\textbf{CommonGen}} & \multicolumn{2}{c}{\textbf{Avg. Gen.}} \\
\cmidrule(lr){2-10}
& \textbf{Acc} $(\uparrow)$ & \textbf{BS} $(\uparrow)$ & \textbf{METEOR} $(\uparrow)$ & \textbf{BS} $(\uparrow)$ & \textbf{METEOR} $(\uparrow)$ & \textbf{BS} $(\uparrow)$ & \textbf{METEOR} $(\uparrow)$ & \textbf{BS} $(\uparrow)$ & \textbf{METEOR} $(\uparrow)$ \\
\midrule

FedAvg & $\textbf{92.77}$ & $41.59$ & $12.37$ & $40.85$ & $13.60$ & $32.28$ & $14.13$ & $38.24$ & $13.37$ \\
FedAvg + Post-train & $\underline{92.73}$ & $42.57$ & $10.07$ & $49.35$ & $19.10$ & $31.20$ & $12.89$ & $\underline{41.04}$ & $\underline{14.02}$ \\
\rowcolor{blue!10}
\textbf{TAP (Ours)} & $91.76$ & $45.07$ & $22.10$ & $52.05$ & $31.08$ & $33.19$ & $13.67$ & $\textbf{43.44}$ & $\textbf{22.28}$ \\
\bottomrule
\end{tabular}%
}
\vspace{-2mm}
\end{table*}

We consider how a network with a large number of clients work with the TAP algorithm. Here, we consider 100 clients with 5 clients participating in aggregation each round on the FLAVA model. The results are presented in Tables \ref{table:appendix-flava-img-100clients} and \ref{table:appendix-txt-100clients}. Table \ref{table:appendix-flava-img-100clients} corresponds to image-based datasets while \ref{table:appendix-txt-100clients} corresponds to text.

Firstly, we can note that the average accuracy and generation scores across both modalities with TAP is generally higher than the baselines. Similar to other results, when TAP is not the best-performing baseline, it is often the next best performing and remains close to the best performer (e.g., $0.5581$ vs $0.5745$ on average image generation MSE score). This indicates that TAP remains robust to settings with large amounts of clients--enabling effective personalization of each individual client in a setting of heterogeneous modalities and tasks on a per-client level.

\subsection{Replacement utilizing Validation Loss}

\begin{table*}[htbp] 
\centering
\caption{FLAVA on image datasets utilizing validation loss for replacement.}
\scriptsize
\resizebox{\linewidth}{!}{%
\label{table:appendix-val-img}
\begin{tabular}{l c c c c c c}
\toprule
\textbf{Method (FLAVA)} & \textbf{Tiny-ImageNet} & \textbf{CIFAR-100} & \textbf{FMNIST} & \textbf{Caltech-256} & \textbf{Avg. Class.} & \textbf{Avg. Gen.} \\
\cmidrule(lr){2-7}
 & \textbf{Acc} $(\uparrow)$ & \textbf{Acc} $(\uparrow)$ & \textbf{MSE} $(\downarrow)$ & \textbf{MSE} $(\downarrow)$ & \textbf{Acc} $(\uparrow)$ & \textbf{MSE} $(\downarrow)$ \\
\midrule

FedAvg & $32.20$ & $48.79$ & $0.6079$ & $0.5100$ & $40.50$ & $0.5590$ \\
FedAvg + Post-train & $40.93$ & $54.26$ & $0.5528$ & $0.4066$ & $47.59$ & $0.4797$ \\
Per-FedAvg & $30.42$ & $44.67$ & $0.7540$ & $0.6579$ & $37.55$ & $0.7060$ \\
Per-FedAvg + Post-train & $41.05$ & $53.43$ & $0.5765$ & $0.4102$ & $47.24$ & $0.4934$ \\
DisentAFL & $36.56$ & $1.23$ & $0.8944$ & $0.5610$ & $18.89$ & $0.7277$ \\
DisentAFL + Post-train & $\underline{46.05}$ & $4.58$ & $0.7650$ & $0.5359$ & $25.31$ & $0.6504$ \\
FedDAT & $15.28$ & $29.82$ & $0.6044$ & $0.4631$ & $22.55$ & $0.5338$ \\
FedDAT + Post-train & $17.84$ & $32.95$ & $0.5504$ & $0.3793$ & $25.39$ & $0.4649$ \\
\rowcolor{blue!10}
\textbf{TAP (train)} & $45.40$ & $\underline{55.47}$ & $\underline{0.5474}$ & $\underline{0.3789}$ & $\underline{50.43}$ & $\underline{0.4632}$ \\
\rowcolor{blue!10}
\textbf{TAP (val)} & $\textbf{46.82}$ & $\textbf{56.07}$ & $\textbf{0.5441}$ & $\textbf{0.3744}$ & $\textbf{51.44}$ & $\textbf{0.4593}$ \\
\bottomrule
\end{tabular}%
}
\end{table*}

\begin{table*}[htbp] 
\centering
\scriptsize
\caption{Performance comparison for FLAVA on text datasets utilizing validation loss for replacement.}
\resizebox{\linewidth}{!}{%
\label{table:appendix-val-txt}
\begin{tabular}{l c cc cc cc cc}
\toprule
\textbf{Method} & \textbf{AG News} & \multicolumn{2}{c}{\textbf{MMLU}} & \multicolumn{2}{c}{\textbf{VQA}} & \multicolumn{2}{c}{\textbf{CommonGen}} & \multicolumn{2}{c}{\textbf{Avg. Gen.}} \\
\cmidrule(lr){2-10}
& \textbf{Acc} $(\uparrow)$ & \textbf{BS} $(\uparrow)$ & \textbf{METEOR} $(\uparrow)$ & \textbf{BS} $(\uparrow)$ & \textbf{METEOR} $(\uparrow)$ & \textbf{BS} $(\uparrow)$ & \textbf{METEOR} $(\uparrow)$ & \textbf{BS} $(\uparrow)$ & \textbf{METEOR} $(\uparrow)$ \\
\midrule

FedAvg & $92.96$ & $39.82$ & $7.12$ & $46.80$ & $18.61$ & $32.47$ & $9.04$ & $39.70$ & $11.59$ \\
FedAvg + Post-train & $\underline{93.07}$ & $40.98$ & $7.09$ & $47.90$ & $19.09$ & $33.84$ & $9.09$ & $40.91$ & $11.76$ \\
Per-FedAvg & $92.71$ & $40.88$ & $9.07$ & $52.40$ & $25.02$ & $32.80$ & $9.46$ & $42.03$ & $14.52$ \\
Per-FedAvg + Post-train & $93.05$ & $39.36$ & $9.39$ & $53.62$ & $25.70$ & $32.82$ & $9.92$ & $41.93$ & $15.00$ \\
DisentAFL & $\textbf{93.11}$ & $39.55$ & $18.77$ & $31.16$ & $5.41$ & $26.50$ & $7.29$ & $32.40$ & $10.49$ \\
DisentAFL + Post-train & $93.00$ & $40.38$ & $20.20$ & $34.81$ & $7.38$ & $26.57$ & $8.58$ & $33.92$ & $12.05$ \\
FedDAT & $92.22$ & $43.78$ & $8.00$ & $62.34$ & $40.77$ & $\underline{39.37}$ & $15.44$ & $48.50$ & $21.40$ \\
FedDAT + Post-train & $92.80$ & $42.97$ & $8.07$ & $61.10$ & $41.08$ & $\textbf{39.58}$ & $\underline{16.23}$ & $47.88$ & $21.79$ \\
\rowcolor{blue!10}
\textbf{TAP (train)} & $92.83$ & $\textbf{50.07}$ & $\underline{22.13}$ & $\textbf{73.68}$ & $\textbf{54.01}$ & $35.66$ & $14.11$ & $\textbf{53.14}$ & $\underline{30.08}$ \\
\rowcolor{blue!10}
\textbf{TAP (val)} & $92.82$ & $\underline{49.01}$ & $\textbf{25.79}$ & $\underline{65.93}$ & $\underline{48.85}$ & $35.65$ & $\textbf{20.22}$ & $\underline{50.20}$ & $\textbf{31.62}$ \\
\bottomrule
\end{tabular}%
}
\vspace{-2mm}
\end{table*}

While the methodology of the TAP algorithm is based primarily on the utilization of the training loss for measuring replacement, we seek to see if utilizing validation loss is also a valid method of training with TAP on FLAVA. For this, we hold-out $10\%$ of each client's dataset as a local validation dataset to use for replacement measurement.

Based on the results in Tables \ref{table:appendix-val-img} and \ref{table:appendix-val-txt}, we can note that the validation loss, in addition to the training loss, are both valid measures of measuring replacement. We can note that with the exception of AG News and CommonGen BS, either val or train with TAP is the best performing method. This indicates that each client can engage in replacement based on either validation or training loss--whichever benefits a certain client more significantly. Overall, this indicates that TAP is flexible to differing loss types in determining whether replacement on the personalized parameters should take place.

\section{Implementation Specifics}\label{appendix:implementation-specs}

\subsection{Decoder Architecture}
\label{appendix:decoder-arch}

Here, we outline the architectures utilized for the decoders $\mathbf{W}^{(D)}$ on differing tasks for both FLAVA and ViLT, presented in Tables \ref{table:classification-head-arch} (classification heads), \ref{table:text-gen-head-arch} (text generation head), and \ref{table:image-generation-head-arch} (image generation head). 

\begin{table}[!ht]
\centering
\caption{Settings of MLP-type classification heads (both text and image).}
\label{table:classification-head-arch}
\begin{tabular}{l c c}
\toprule
\textbf{Type} & \textbf{FLAVA}~\citep{singh2022flava} & \textbf{ViLT}~\citep{kim2021vilt} \\
\midrule
Num. of Linear Layers & 2 & 3 \\
Activation Func. & ReLU & GELU \\
Hidden Dimension & embed\_dim / 2 & [embed\_dim * 4, embed\_dim] \\
Dropout & 0.3 & 0.1 \\
LayerNorm Utilized & No & Yes \\
\bottomrule
\end{tabular}%
\end{table}

\begin{table}[H] 
\centering
\caption{Architecture of the text generation head for both FLAVA and ViLT.}
\label{table:text-gen-head-arch}
\begin{tabular}{l c}
\toprule
\textbf{Component} & \textbf{Specification} \\
\midrule
Head Type & Conditional Autoregressive (GPT-2 based) \\
\midrule
\textbf{Conditioning Mechanism} & \\
\quad Layer Type & Multi-Head Attention \\
\quad \# of Attention Heads & 8 \\
\quad Embedding Dimension & 768 \\
\midrule
\textbf{Core Generator} & \\
\quad Architecture & GPT-2 \\
\quad \# of Layers & 1 \\
\quad \# of Attention Heads & 4 \\
\quad Hidden Dimension & 768 \\
\quad Inner FFN Dimension & 1536 (2 $\times$ 768) \\
\midrule
\textbf{Output Projection} & \\
\quad Layer Type & Linear (no bias) \\
\bottomrule
\end{tabular}%
\end{table}

\begin{table}[H] 
\centering
\caption{Architecture of the image generation head used in both FLAVA and ViLT.}
\label{table:image-generation-head-arch}
\begin{tabular}{l c}
\toprule
\textbf{Component} & \textbf{Specification} \\
\midrule
Head Type & Transposed Convolutional Decoder \\
Input Embedding Dimension & 768 \\
Target Image Size & 64×64 \\
Output Channels & 3 (RGB) \\
\midrule
\textbf{Initial Projection} & \\
\quad Layer Type & Linear \\
\quad Output Dimensions & 64 × 8 × 8 \\
\midrule
\textbf{Decoder Layers} & \\
\quad Layer 1: ConvTranspose2d & \\
\quad \quad Input/Output Channels & 64 → 32 \\
\quad \quad Kernel/Stride/Padding & 4/2/1 \\
\quad \quad Output Resolution & 16×16 \\
\quad Activation & ReLU \\
\quad Layer 2: ConvTranspose2d & \\
\quad \quad Input/Output Channels & 32 → 16 \\
\quad \quad Kernel/Stride/Padding & 4/2/1 \\
\quad \quad Output Resolution & 32×32 \\
\quad Activation & ReLU \\
\quad Layer 3: ConvTranspose2d & \\
\quad \quad Input/Output Channels & 16 → 3 \\
\quad \quad Kernel/Stride/Padding & 4/2/1 \\
\quad \quad Output Resolution & 64×64 \\
\bottomrule
\end{tabular}%
\end{table}

\subsection{Hyperparameter Setup}
\label{appendix:hyperparameters}

In Table \ref{table:hyperparams-setup}, general hyperparameters utilized in running our experiments are outlined below. Table \ref{table:hyperparams-setup-margin} details the margin values $m_{i,o}$ utilized in the TAP algorithm to determine whether replacement will take place (Stage 1 of Fig.~\ref{fig:two-stage_method}). Unless stated otherwise, the following are the settings across all experiments.

\begin{table}[!ht]
\centering
\caption{Hyperparameter settings to run numerical experiments.}
\label{table:hyperparams-setup}
\begin{tabular}{l c c}
\toprule
\textbf{Hyperparameters} & \textbf{FLAVA}~\citep{singh2022flava} & \textbf{ViLT}~\citep{kim2021vilt} \\
\midrule
Num. of Clients Aggregated each Round & $2$ & $2$ \\
Batch Size & $128$ & $128$ \\
LoRA Encoder Attention Rank & $8$ & $4$ \\
LoRA Backbone Attention Rank & $16$ & $8$ \\
LoRA Backbone Expert Rank & $4$ & $4$ \\
LoRA Dropout Rate & $0.3$ & $0.3$ \\
FedDAT~\citep{chen2024feddat} Adapter Bottleneck Size & $32$ & $32$ \\
KD Temperature $\widetilde{\tau}$ & $1$ & $1$ \\
Disent. Loss Weight & $0.5$ & $0.5$ \\
Image Gen. Distill. Weight & 2e-3 & 2e-3 \\
Text Gen. Distill. Weight & 1 & 1 \\
Image Classification Distill. Weight & 2e-3 & 2e-3 \\
Text Classification Distill. Weight & 2e-3 & 2e-3 \\
FedDAT KL Weight, $\forall o \in \mathcal{O}$ & 2e-3 & 2e-3 \\
Initial Learning Rate $\eta_0$ & 1e-4 & 1e-5 \\
Post-warmup Learning Rate $\eta_t$ & 3e-4 & 5e-5 \\
Per-FedAvg~\citep{fallah2020personalized} inner update initial learning rate & 1e-4 & 1e-5 \\
Per-FedAvg post-warmup inner update learning rate & 3e-4 & 5e-5 \\
AdamW weight decay & $0.01$ & $0.01$ \\
Num. of Warmup Rounds & $20$ & $20$ \\
Num. Local Iterations & $20$ & $20$ \\
Num. Distillation Iterations $P$ & $50$ & $50$ \\
Total Communication Rounds $T$ & $200$ & $200$ \\
\bottomrule
\end{tabular}%
\end{table}

\begin{table}[!ht] 
\centering
\caption{Margin settings to run numerical experiments.}
\label{table:hyperparams-setup-margin}
\begin{tabular}{l c c}
\toprule
\textbf{Task Type} & \textbf{FLAVA}~\citep{singh2022flava} & \textbf{ViLT}~\citep{kim2021vilt} \\
\midrule
Image Classification Task(s) & $0.01$ & --- \\
Image Generation Task(s) & $0.005$ & $0.005$ \\
Text Classification Task(s) & $0.005$ & $0.005$ \\
Text Generation Task(s) & $0.01$ & $0.01$ \\
\bottomrule
\end{tabular}%
\end{table}

\subsection{Text Templates}
\label{appendix:templates}

In Table \ref{table:text-templates}, we specify the template format utilized for each text generation-based dataset (MMLU, VQA, and CommonGen). Portions surrounded with $\{ \}$ brackets indicate where certain data fields are to be inputted.

\begin{table}[H]
\centering
\caption{Templates utilized for the text generation datasets. The inputs are blue, with the model's expected response marked in red.}
\scriptsize
\label{table:text-templates}
\begin{tabular}{@{}>{\centering\arraybackslash}p{3cm}|p{10cm}@{}}
\toprule
\textbf{Dataset} & \textbf{Template} \\ 
\midrule

MMLU &
\tcbox[
  colback=gray!10,
  colframe=black,
  boxsep=1pt,
  top=1pt,
  bottom=1pt,
  left=2pt,
  right=2pt]{
\begin{minipage}[t]{0.92\linewidth}
\textcolor{blue}{The following is a multiple-choice question about \{SUBJECT\}}.  \\
\textcolor{blue}{Choose the correct answer from the options below.} \\
\textcolor{blue}{Question: \{QUESTION\} \\
Options: \\
\{CHOICES\}}\\
\\
\textcolor{red}{Correct answer on \{SUBJECT\}: \{ANSWER\}}
\end{minipage}
} \\ \midrule

VQA &
\tcbox[colback=gray!10,colframe=black]{
\begin{minipage}[t]{0.85\linewidth}
\textcolor{blue}{Answer the question based on the image. \\
Question: \{QUESTION\}} \\
\\
\textcolor{red}{Answer: \{ANSWER\}}
\end{minipage}
} \\ \midrule

CommonGen &
\tcbox[colback=gray!10,colframe=black]{
\begin{minipage}[t]{0.85\linewidth}
\textcolor{blue}{Generate an appropriate description from the concepts below. \\
Concepts: \{CONCEPTS\}} \\
\\
\textcolor{red}{Description: \{ANSWER\}}
\end{minipage}
} \\

\bottomrule
\end{tabular}%
\end{table}

\section{Analysis}\label{appendix:proof-derivation}

\subsection{Lemmas}
As a preliminary, we introduce some basic lemmas, which will be utilized in the derivation of the theorem.
\begin{lemma}\label{lemma:cauchy}
    For any vectors $v_1, \ldots, v_N$, we have
    \[
    \left| \left| \sum_{n=1}^{N} v_n \right| \right|^2 \leq N \sum_{n=1}^{N} \| v_n \|^2.
    \]
\end{lemma}
\begin{lemma}\label{lemma:orthogonal}
    If vectors $v_1, \ldots, v_N$ are such that $v_n$ only has non-zero values on a unique block of coordinates (e.g. $\langle v_n, v_{n'} \rangle = 0$ for $n \neq n'$), then
    \[
    \left| \left| \sum_{n=1}^{N} v_n \right| \right|^2 = \sum_{n=1}^{N} \| v_n \|^2.
    \]
\end{lemma}

\subsubsection{Proof of Lemma \ref{lemma:cauchy}}
Given vectors $v_1, \ldots, v_N $, we have
\[
\left| \left| \sum_{n=1}^{N} v_n \right| \right|^2 = \langle \sum_{n=1}^{N} v_n,  \sum_{n'=1}^{N} v_{n'} \rangle =  \sum_{n=1}^{N} \sum_{n'=1}^{N} \langle v_n, v_{n'} \rangle,
\]
which can be decomposed and expressed via the following:
\[
\sum_{n=1}^{N} \| v_n \|^2 + 2\sum_{1 \leq n < n' \leq N} \langle v_n, v_{n'} \rangle.
\]

Then, by Cauchy-Schwarz, where $\langle v_n, v_{n'} \rangle \leq \|v_n\|\|v_{n'}\|$, $n \neq n'$, we have
\[
\left| \left| \sum_{n=1}^{N} v_n \right| \right|^2 \leq \sum_{n=1}^{N} \| v_n \|^2 + 2\sum_{1 \leq n < n' \leq N} \|v_n\|\|v_{n'}\|,
\]

where the RHS is equivalent to $\left( \sum_{n=1}^{N} \| v_n \| \right)^2$. Then by using Cauchy-Schwarz again, we obtain
\[
\left| \left| \sum_{n=1}^{N} v_n \right| \right|^2 \leq \left( \sum_{n=1}^{N} \| v_n \| \right)^2 \leq \left( \sum_{n=1}^{N} 1^2 \right) \left( \sum_{n=1}^{N} \| v_n \|^2 \right),
\]
which gives
\[
\left| \left| \sum_{n=1}^{N} v_n \right| \right|^2 \leq  N \left( \sum_{n=1}^{N} \| v_n \|^2 \right).
\]

\subsubsection{Proof of Lemma \ref{lemma:orthogonal}}
Given vectors $v_1, \ldots, v_N $, we have
\[
\left| \left| \sum_{n=1}^{N} v_n \right| \right|^2 = \langle \sum_{n=1}^{N} v_n,  \sum_{n'=1}^{N} v_{n'} \rangle =  \sum_{n=1}^{N} \sum_{n'=1}^{N} \langle v_n, v_{n'} \rangle.
\]

Due to orthogonality, $\langle v_n, v_{n'} \rangle = 0$ for $n \neq n'$, and with $n=n'$ being $\| v_n \|^2$, we have
\[
\left| \left| \sum_{n=1}^{N} v_n \right| \right|^2 = \sum_{n=1}^{N} \| v_n \|^2,
\]
which completes the proof.

\subsection{Proof of Theorem \ref{theorem:server-bound}}\label{proof:server-bound}
Firstly, from Assumption \ref{assump:l-smooth}, we have:
\begin{equation}\label{eq:personal-smooth}
    f(\widetilde{\mathbf{W}}_{t+1}) \leq f(\widetilde{\mathbf{W}}_{t}) + \langle \nabla f(\widetilde{\mathbf{W}}_{t}), \widetilde{\mathbf{W}}_{t+1} - \widetilde{\mathbf{W}}_{t}\rangle + \frac{L}{2} \| \widetilde{\mathbf{W}}_{t+1} - \widetilde{\mathbf{W}}_{t} \|^2.
\end{equation}
Then, decompose the update by blocks and take expectation over round-$t$ randomness:
\begin{equation}
    \mathbb{E} \left[ f(\widetilde{\mathbf{W}}_{t+1}) \right] \leq f(\widetilde{\mathbf{W}}_{t}) + \sum_{r=0}^{R} \mathbb{E} \langle\nabla f(\widetilde{\mathbf{W}}_{t, [r]}), \Delta_{t, [r]}\rangle + \frac{L}{2} \mathbb{E} \left| \left| \sum_{r=0}^{R} \Delta_{t, [r]} \right| \right|^2,
\end{equation}
where $\Delta_{t, [r]} = -\frac{\eta_t}{K_r}\sum_{i: \mathcal{B}_r \in \mathcal{B}_i} \sum_{s=0}^{\tau - 1} \nabla \tilde{\ell} \left(\mathcal{H}_i; P_{\mathcal{B}_r}\widetilde{\mathbf{W}}_{t, [i]}^{(s)}\right)$, with $\tau$ as the number of local training iterations. Therefore, $\widetilde{\mathbf{W}}_{t+1} = \widetilde{\mathbf{W}}_t + \sum_{r=0}^{R} \Delta_{t, [r]}$. Then, by Assumption \ref{assump:bounded-variance}, $\mathbb{E} \langle\nabla f(\widetilde{\mathbf{W}}_{t, [r]}), \Delta_{t, [r]}\rangle =  -\eta_t \sum_{s=0}^{\tau - 1} \mathbb{E} \langle \nabla f(\widetilde{\mathbf{W}}_{t, [r]}), \frac{1}{K_r} \sum_{i: \mathcal{B}_r \in \mathcal{B}_i} \nabla g_i (\widetilde{\mathbf{W}}_{t, [r]}^{(s)}) \rangle$. Next, with $- \langle a,b \rangle = \frac{1}{2} \| a-b \|^2 - \frac{1}{2} \|a \|^2 - \frac{1}{2} \| b \|^2$, where $a = \nabla f (\widetilde{\mathbf{W}}_{t, [r]})$ and $b = \frac{1}{K_r} \sum_{i: \mathcal{B}_r \in \mathcal{B}_i} \nabla g_i (\widetilde{\mathbf{W}}_{t, [r]}^{(s)})$, we can write the second term on the RHS as 
\begin{align}
    \mathbb{E} \left[ f(\widetilde{\mathbf{W}}_{t+1}) \right] \leq & 
    f(\widetilde{\mathbf{W}}_{t}) + \sum_{r=0}^{R} \frac{\eta_t}{2} \sum_{s=0}^{\tau - 1} \mathbb{E} \left| \left| \nabla f (\widetilde{\mathbf{W}}_{t, [r]}) - \frac{1}{K_r} \sum_{i: \mathcal{B}_r \in \mathcal{B}_i} \nabla g_i (\widetilde{\mathbf{W}}_{t, [r]}^{(s)}) \right| \right|^2 \nonumber \\
    & -  \sum_{r=0}^{R}\frac{\eta_t \tau}{2} \| f (\widetilde{\mathbf{W}}_{t, [r]}) \|^2 - \sum_{r=0}^{R} \frac{\eta_t}{2} \sum_{s=0}^{\tau - 1} \mathbb{E} \left| \left| \frac{1}{K_r} \sum_{i: \mathcal{B}_r \in \mathcal{B}_i} \nabla g_i (\widetilde{\mathbf{W}}_{t, [r]}^{(s)}) \right| \right|^2  \nonumber \\
    & + \frac{L}{2} \mathbb{E} \left| \left| \sum_{r=0}^{R} \Delta_{t, [r]} \right| \right|^2.
\end{align}

Then, discard the second to last term on the RHS. For the second term of the RHS, using the fact $\nabla f (\widetilde{\mathbf{W}}_{[r]}) = \frac{1}{K_r} \sum_{i: \mathcal{B}_r \in \mathcal{B}_i} \nabla g_i (\widetilde{\mathbf{W}}_{[r]})$, Jensen's inequality, and Assumption \ref{assump:l-smooth}, we derive
\begin{align}
    \mathbb{E} \left[ f(\widetilde{\mathbf{W}}_{t+1}) \right] \leq 
    f(\widetilde{\mathbf{W}}_{t}) - \frac{\eta_t \tau}{2} \sum_{r=0}^{R} \| \nabla f(\widetilde{\mathbf{W}}_{t, [r]}) \|^2 + \frac{\eta_t L^2}{2} \sum_{r=0}^{R} \sum_{s=0}^{\tau - 1} \psi_{t,r}^{(s)} + \frac{L}{2} \mathbb{E} \left| \left| \sum_{r=0}^{R} \Delta_{t, [r]} \right| \right|^2,
\end{align}
with $\psi_{t, r}^{(s)} := \frac{1}{K_r} \sum_{i: \mathcal{B}_r \in \mathcal{B}_i} \mathbb{E} \| \widetilde{\mathbf{W}}_{t, [i]}^{(s)} - \widetilde{\mathbf{W}}_{t, [r]} \|^2$. Then for the last term of the RHS, using the property of $\| u + v \|^2 \leq 2\| u \|^2 + 2\| v \|^2$ from Lemma \ref{lemma:cauchy}, we can obtain
\begin{align}
    &\mathbb{E} \left[ f(\widetilde{\mathbf{W}}_{t+1}) \right] \leq 
    f(\widetilde{\mathbf{W}}_{t}) - \frac{\eta_t \tau}{2} \sum_{r=0}^{R} \| \nabla f(\widetilde{\mathbf{W}}_{t, [r]}) \|^2 + \frac{\eta_t L^2}{2} \sum_{r=0}^{R} \sum_{s=0}^{\tau - 1} \psi_{t,r}^{(s)} \nonumber \\
    &+ L \sum_{r=0}^{R=0} \frac{\eta_t^2}{K_r^2}\left( 2 \mathbb{E} \left| \left| \sum_{i: \mathcal{B}_r \in \mathcal{B}_i} \sum_{s=0}^{\tau - 1} \left( \nabla \widetilde{\ell} (\cdot) - \nabla g_i (\widetilde{\mathbf{W}}_{t}^{(s)}) \right) \right| \right|^2 + 2 \mathbb{E} \left| \left| \sum_{i: \mathcal{B}_r \in \mathcal{B}_i} \sum_{s=0}^{\tau - 1} \nabla g_i (\widetilde{\mathbf{W}}_{t}^{(s)}) \right| \right|^2 \right),
\end{align}
where we abbreviate $\nabla \widetilde{\ell}(\cdot) = \nabla \widetilde{\ell}\left( \mathcal{H}_i; P_{\mathcal{B}_r} \widetilde{\mathbf{W}}_{t, [i]}^{(s)} \right)$. This can be further simplified with the variance term by Assumption \ref{assump:bounded-variance} via
\begin{align}
    \mathbb{E} \left[ f(\widetilde{\mathbf{W}}_{t+1}) \right] \leq &
    f(\widetilde{\mathbf{W}}_{t}) - \frac{\eta_t \tau}{2} \sum_{r=0}^{R}  \| \nabla f(\widetilde{\mathbf{W}}_{t, [r]}) \|^2 + \frac{\eta_t L^2}{2} \sum_{r=0}^{R} \sum_{s=0}^{\tau - 1} \psi_{t,r}^{(s)} \nonumber \\
    &+ L \sum_{r=0}^{R=0} \frac{\eta_t^2}{K_r^2}\left( 2 K_r \tau \sigma^2 + 2 \mathbb{E} \left| \left| \sum_{i: \mathcal{B}_r \in \mathcal{B}_i} \sum_{s=0}^{\tau - 1} \nabla g_i (\widetilde{\mathbf{W}}_{t}^{(s)}) \right| \right|^2 \right).
\end{align}
Moreover, via Lemma \ref{lemma:cauchy}, we can further say
\begin{align}
    \mathbb{E} \left[ f(\widetilde{\mathbf{W}}_{t+1}) \right] \leq &
    f(\widetilde{\mathbf{W}}_{t}) - \frac{\eta_t \tau}{2} \sum_{r=0}^{R} \| \nabla f(\widetilde{\mathbf{W}}_{t, [r]}) \|^2 + \frac{\eta_t L^2}{2} \sum_{r=0}^{R} \sum_{s=0}^{\tau - 1} \psi_{t,r}^{(s)} \nonumber \\
    &+ L \sum_{r=0}^{R=0} \frac{\eta_t^2}{K_r^2}\left( 2 K_r \tau \sigma^2 + 2 K_r \tau \sum_{i: \mathcal{B}_r \in \mathcal{B}_i} \sum_{s=0}^{\tau - 1} \mathbb{E} \| \nabla g_i (\widetilde{\mathbf{W}}_{t}^{(s)}) \|^2 \right)
\end{align}
\begin{align}\label{proof:before-split}
    \mathbb{E} \left[ f(\widetilde{\mathbf{W}}_{t+1}) \right] \leq &
    f(\widetilde{\mathbf{W}}_{t}) - \frac{\eta_t \tau}{2} \sum_{r=0}^{R} \| \nabla f(\widetilde{\mathbf{W}}_{t, [r]}) \|^2 + \frac{\eta_t L^2}{2} \sum_{r=0}^{R} \sum_{s=0}^{\tau - 1} \psi_{t,r}^{(s)} \nonumber \\
    &+ L \sum_{r=0}^{R=0} \frac{\eta_t^2}{K_r}\left( 2 \tau \sigma^2 + 2 \tau \sum_{i: \mathcal{B}_r \in \mathcal{B}_i} \sum_{s=0}^{\tau - 1}  \mathbb{E} \| \nabla g_i (\widetilde{\mathbf{W}}_{t}^{(s)}) \|^2 \right).
\end{align}

For the last term of the RHS, split the parentheses into
\[
2L\eta_t^2\tau \sum_{r=0}^{R} \frac{\sigma^2}{K_r} + 2L \eta_t^2\tau \sum_{r=0}^{R} \sum_{s=0}^{\tau - 1} \left( \frac{1}{K_r} \sum_{i: \mathcal{B}_r \in \mathcal{B}_i} \mathbb{E} \| \nabla g_i (\widetilde{\mathbf{W}}_t^{(s)}) \|^2 \right),
\]
and then decompose 
\[
\nabla g_i(\widetilde{\mathbf{W}}^{(s)}_{t})
=\underbrace{\nabla f(\widetilde{\mathbf{W}}_{t, [r]})}_{u}
+\underbrace{\big(\nabla g_i(\widetilde{\mathbf{W}}_{t, [r]})-\nabla f(\widetilde{\mathbf{W}}_{t, [r]})\big)}_{v}
+\underbrace{\big(\nabla g_i(\widetilde{\mathbf{W}}^{(s)}_{t})-\nabla g_i(\widetilde{\mathbf{W}}_{t, [r]})\big)}_{w}.
\]

Next, using $\|u+v+w\|^2\le 3(\|u\|^2+\|v\|^2+\|w\|^2)$ from Lemma \ref{lemma:cauchy} in conjunction with Assumptions \ref{assump:l-smooth} and \ref{assump:bounded-heterogeneity}, we have
\[
\frac{1}{K_r} \sum_{i: \mathcal{B}_r \in \mathcal{B}_i} \mathbb{E} \| \nabla g_i(\widetilde{\mathbf{W}}_t^{(s)}) \|^2 \leq 3\|\nabla f(\widetilde{\mathbf{W}}_{t, [r]})\|^2 + 3\zeta_r^2 + 3L^2\psi_{t,r}^{(s)}.
\]

Then, summing over $s=0, \ldots, \tau-1$, we get
\[
\sum_{s=0}^{\tau-1} \frac{1}{K_r} \sum_{i: \mathcal{B}_r \in \mathcal{B}_i} \mathbb{E} \| \nabla g_i(\widetilde{\mathbf{W}}_t^{(s)}) \|^2 \leq \tau (3\|\nabla f(\widetilde{\mathbf{W}}_{t, [r]})\|^2 + 3\zeta_r^2) + 3L^2\sum_{s=0}^{\tau - 1}\psi_{t,r}^{(s)}.
\]

Now plug the above back into \eqref{proof:before-split} and use Lemma \ref{lemma:orthogonal} to get
\begin{align}\label{proof:before-psi-unroll}
    \mathbb{E} \left[ f(\widetilde{\mathbf{W}}_{t+1}) \right] \leq &
    f(\widetilde{\mathbf{W}}_{t}) - \left( \frac{\eta_t \tau}{2} - 6L\eta_t^2\tau^2 \right) \| \nabla f(\widetilde{\mathbf{W}}_t) \|^2 + \left( \frac{\eta_t L^2}{2} + 6L^3\eta_t^2\tau \right) \sum_{r=0}^{R} \sum_{s=0}^{\tau-1}\psi_{t,r}^{(s)} \nonumber \\
    & + 6L\eta_t^2\tau^2\sum_{r=0}^{R}\zeta_r^2 + 2L\eta_t^2\tau \sum_{r=0}^{R} \frac{\sigma^2}{K_r}.
\end{align}

Then, for bounding the summations with $\psi_{t, r}^{(s)}$, firstly note the fact that $\widetilde{\mathbf{W}}_{t, [i]}^{(s+1)} - \widetilde{\mathbf{W}}_{t, [r]} = \widetilde{\mathbf{W}}_{t, [i]}^{(s)} - \widetilde{\mathbf{W}}_{t, [r]} - \eta_t \left( \nabla g_i (\widetilde{\mathbf{W}}_{t}^{(s)}) + \bm\xi^{(s)}_{t,i} \right)$, where $\bm\xi^{(s)}_{t,i} := \nabla \widetilde{\ell}(\cdot) - \nabla g_i (\widetilde{\mathbf{W}}_{t}^{(s)})$, with zero mean and $\mathbb{E} \|\bm\xi^{(s)}_{t,i} \|^2 \leq \sigma^2$. Therefore,
\begin{equation}
    \begin{aligned}
        \mathbb{E} \| \widetilde{\mathbf{W}}_{t, [i]}^{(s)} - \widetilde{\mathbf{W}}_{t, [r]} \|^2 \leq &  \mathbb{E} \| \sum_{h=0}^{s-1}-\eta_t(\nabla g_i (\widetilde{\mathbf{W}}_{t}^{(h)}) + \bm\xi^{(h)}_{t,i} ) \|^2  \\
        \leq & s \sum_{h=0}^{s-1} \eta_t^2 \mathbb{E} \| \nabla g_i (\widetilde{\mathbf{W}}_{t}^{(h)}) + \bm\xi^{(h)}_{t,i} \|^2  \\
        \leq & \eta_t^2 s \sum_{h=0}^{s-1} \mathbb{E} \| \nabla g_i (\widetilde{\mathbf{W}}_{t}^{(h)}) \|^2 + \eta_t^2 s \sum_{h=0}^{s-1}\mathbb{E} \| \bm\xi^{(h)}_{t,i} \|^2.
    \end{aligned}
\end{equation}

Averaging over clients and recalling the definition of $\psi_{t,r}^{(s)}$, we can derive 
\[
\begin{aligned}
&\psi_{t,r}^{(s)} \leq \eta_t^2 s \sum_{h=0}^{s - 1} \left( \frac{1}{K_r} \sum_{i: \mathcal{B}_r \in \mathcal{B}_i} \mathbb{E} \| \nabla g_i (\widetilde{\mathbf{W}}_t^{(h)}) \|^2 \right) + \eta_t^2 s \sum_{h=0}^{s-1}\left( \frac{1}{K_r} \sum_{i: \mathcal{B}_r \in \mathcal{B}_i} \mathbb{E} \| \bm\xi^{(h)}_{t,i} \|^2 \right) \\
& \psi_{t,r}^{(s)} \leq \eta_t^2 s \sum_{h=0}^{s-1} \left( 3 \| \nabla f(\widetilde{\mathbf{W}}_{t, [r]}) \|^2 + 3\zeta_r^2 + 3L^2\psi_{t,r}^{(h)} \right) + \eta_t^2 s^2 \sigma^2.
\end{aligned}
\]
Summing them from $s=0$ to $\tau-1$ gives rise to 
\[
\Psi_{t,r} \leq  3 \eta_t^2 \sum_{s=0}^{\tau-1} s \sum_{h=0}^{s-1} \| \nabla f(\widetilde{\mathbf{W}}_{t, [r]}) \|^2 + 3\eta_t^2 \sum_{s=0}^{\tau-1} s \sum_{h=0}^{s-1}\zeta_r^2 + 3L^2 \eta_t^2 \sum_{s=0}^{\tau-1} s \sum_{h=0}^{s-1} \psi_{t,r}^{(h)}  + \frac{1}{3} \eta_t^2 {\tau}^3 \sigma^2,
\]
where we used the fact that $\sum_{s=0}^{\tau-1} s^2 \leq \frac{1}{3}H^3$. We can further simplify it as
    \begin{align}
        \Psi_{t,r} \leq &  \eta_t^2 \tau^3 \| \nabla f(\widetilde{\mathbf{W}}_{t, [r]}) \|^2 + 2 \eta_t^2 \tau^2 L^2 \sum_{s=0}^{\tau-1} \psi_{t,r}^{(s)} + \eta_t^2 {\tau}^3 \zeta_r^2  + \frac{1}{3} \eta_t^2 {\tau}^3 \sigma^2 \nonumber \\
        \leq &  \eta_t^2 \tau^3 \| \nabla f(\widetilde{\mathbf{W}}_{t, [r]}) \|^2 + 2 \eta_t^2 \tau^2 L^2 \Psi_{t,r} + \eta_t^2 {\tau}^3 \zeta_r^2  + \frac{1}{3} \eta_t^2 {\tau}^3 \sigma^2.
    \end{align}
Reorganizing it gives rise to 
\[
(1- 2 \eta_t^2 \tau^2 L^2)\Psi_{t,r} \leq \eta_t^2 \tau^3 \| \nabla f(\widetilde{\mathbf{W}}_{t, [r]}) \|^2  + \eta_t^2 {\tau}^3 \zeta_r^2  + \frac{1}{3} \eta_t^2 {\tau}^3 \sigma^2.
\]

Then, setting $\eta_t \leq \frac{1}{2L \tau}$ such that $1-2L^2\eta_t^2\tau^2 \geq \frac{1}{2}$, we have $\Psi_{t,r} \leq 2\eta_t^2\tau^3 \left( \| \nabla f(\widetilde{\mathbf{W}}_{t,[r]}) \|^2 + \zeta_r^2 + \frac{1}{3}\sigma^2 \right)$. Summing over $r$ and utilizing Lemma \ref{lemma:orthogonal}, we derive
\[
\sum_{r=0}^{R} \Psi_{t,r} \leq 2\eta_t^2\tau^3 \left( \| \nabla f(\widetilde{\mathbf{W}}_t) \|^2 + \sum_{r=0}^{R} \zeta_r^2 + \frac{R+1}{3}\sigma^2 \right),
\]
which when plugged back into \eqref{proof:before-psi-unroll} gives rise to
\begin{align}
    \mathbb{E} \left[ f(\widetilde{\mathbf{W}}_{t+1}) \right] \leq &
    f(\widetilde{\mathbf{W}}_{t}) - \left( \frac{\eta_t \tau}{2} - 6L\eta_t^2\tau^2 \right) \| \nabla f(\widetilde{\mathbf{W}}_t) \|^2 \nonumber \\
    & + 2\eta_t^2\tau^3\left( \frac{\eta_t L^2}{2} + 6L^3\eta_t^2\tau \right) \left( \| \nabla f(\widetilde{\mathbf{W}}_t) \|^2 + Z + \frac{R+1}{3}\sigma^2 \right) \nonumber \\
    & + 6L\eta_t^2\tau^2 Z + 2L\eta_t^2\tau\sigma^2C_{K},
\end{align}
where $Z := \sum_{r=0}^{R}\zeta_r^2$ and $C_K := \sum_{r=0}^{R} \frac{1}{K_r}$. Then group the terms in $\| \nabla f(\widetilde{\mathbf{W}}_t) \|^2$ to get $\eta_t \tau \left( \frac{1}{2} - 6L\eta_t\tau - L^2\eta_t^2\tau^2-12L^3\eta_t^3\tau^3 \right)$. Next, by setting the learning rate as $\eta_t \leq \min \left\{ \frac{1}{48L\tau}, \frac{1}{\sqrt{8}L\tau}, \left( \frac{1}{96L^3\tau^3} \right)^{\frac{1}{3}} \right\}$, we give rise to
\begin{align}
    \frac{\eta_t }{8} \| \nabla f(\widetilde{\mathbf{W}}_t) \|^2 \leq &
    f(\widetilde{\mathbf{W}}_{t}) - \mathbb{E} \left[ f(\widetilde{\mathbf{W}}_{t+1}) \right] \nonumber \\
    & + 2\eta_t^2\tau^3\left( \frac{\eta_t L^2}{2} + 6L^3\eta_t^2\tau \right) \left( Z + \frac{R+1}{3}\sigma^2 \right) \nonumber \\
    & + 6L\eta_t^2\tau^2 Z + 2L\eta_t^2\tau\sigma^2C_{K}.
\end{align}

Moreover, enforcing $6L^3\eta_t^2\tau \leq \frac{\eta_t L^2}{2}$ (when $\eta_t \leq \frac{1}{12L\tau}$, which is weaker than $\eta_t \leq \frac{1}{48L\tau}$), we can further derive 
\begin{align}
    \frac{\eta_t }{8}\| \nabla f(\widetilde{\mathbf{W}}_t) \|^2 \leq &
    f(\widetilde{\mathbf{W}}_{t}) - \mathbb{E} \left[ f(\widetilde{\mathbf{W}}_{t+1}) \right] + 2\eta_t^2\tau^3\left(\eta_t L^2\right) \left( Z + \frac{R+1}{3}\sigma^2 \right) \nonumber \\
    & + 6L\eta_t^2\tau^2 Z + 2L\eta_t^2\tau\sigma^2C_{K}.
\end{align}

Now taking expectation and summing them from $t=0$ to $T-1$, we obtain
\begin{align}\label{bound_proof_21}
    \frac{1}{\sum_{t=0}^{T-1} \eta_t }\sum_{t=0}^{T-1} \eta_t \mathbb{E} \| \nabla f(\widetilde{\mathbf{W}}_t) \|^2 \leq & \frac{8 \left( f(\widetilde{\mathbf{W}}_0) - \mathbb{E}\left[f(\widetilde{\mathbf{W}}_T)\right] \right)}{\sum_{t=0}^{T-1} \eta_t} +16 \left(Z+\frac{R+1}{3}\sigma^2 \right) \frac{\tau^3L^2}{\sum_{t=0}^{T-1} \eta_t}\sum_{t=0}^{T-1} \eta^3_t \nonumber \\
    & + 48L \tau^2 Z \frac{1}{\sum_{t=0}^{T-1} \eta_t}\sum_{t=0}^{T-1} \eta_t^2 + 16 L \tau \sigma^2C_{K} \frac{1}{\sum_{t=0}^{T-1} \eta_t}\sum_{t=0}^{T-1} \eta_t^2.
\end{align}
Note that $R\geq1$. This thus completes the proof.

\end{document}